\documentclass[11pt,a4paper]{article}
\usepackage[hyperref]{acl2020}
\usepackage{times}
\usepackage{latexsym}
\usepackage{array}
\usepackage{url}
\usepackage{float}
\usepackage{graphicx}
\usepackage{booktabs}
\usepackage{tabularx,paralist}
\usepackage{amssymb}
\usepackage{algorithm}
\usepackage{algpseudocode}
\usepackage{microtype}
\usepackage{xcolor}
\usepackage{stfloats}

\aclfinalcopy % Uncomment this line for the final submission
\newcommand{\SideNote}[2]{} 
\renewcommand{\SideNote}[2]{\todo[color=#1,size=\footnotesize]{#2}}

\title{On the Robustness of Language Encoders against Grammatical Errors}

\author{Fan Yin$^1$, Quanyu Long$^2$, Tao Meng$^3$, and Kai-Wei Chang$^3$ \\
  $^1$Peking University \\
  $^2$Shanghai Jiao Tong University\\
  $^3$University of California, Los Angeles \\
  \texttt{1600012975@pku.edu.cn};  \\
  \texttt{oscar.long@sjtu.edu.cn}; \\
  \texttt{tmeng@cs.ucla.edu}; \\
  \texttt{kwchang@cs.ucla.edu}
  }

\date{}

\begin{document}
\maketitle
\begin{abstract}
We conduct a thorough study to diagnose the behaviors of pre-trained language encoders (ELMo, BERT, and RoBERTa) when confronted with natural grammatical errors. Specifically, we collect real grammatical errors from non-native speakers and conduct adversarial attacks to simulate these errors on clean text data. We use this approach to facilitate debugging models on downstream applications. Results confirm that the performance of all tested models is affected but the degree of impact varies. To interpret model behaviors, we further design a linguistic acceptability task to reveal their abilities in identifying ungrammatical sentences and the position of errors. We find that fixed contextual encoders with a simple classifier trained on the prediction of sentence correctness are able to locate error positions. We also design a cloze test for BERT and discover that BERT captures the interaction between errors and specific tokens in context. Our results shed light on understanding the robustness and behaviors of language encoders against grammatical errors. 

\end{abstract}

\section{Introduction}

Pre-trained language encoders have achieved great success in facilitating various downstream natural language processing (NLP) tasks \citep{DBLP:conf/naacl/PetersNIGCLZ18, DBLP:conf/naacl/DevlinCLT19, DBLP:journals/corr/abs-1907-11692}. However, they usually assume training and test corpora are clean and it is unclear how the models behave when confronted with noisy input. Grammatical error is an important type of noise since it naturally and frequently occurs in natural language, especially in spoken and written materials from non-native speakers. Dealing with such a noise reflects model robustness in representing language and grammatical knowledge. It would also have a positive social impact if language encoders can model texts from non-native speakers appropriately.

Recent work on evaluating model's behaviors against grammatical errors employs various methods, including (1) manually constructing minimal edited pairs on specific linguistic phenomena \citep{DBLP:conf/emnlp/MarvinL18,DBLP:journals/corr/abs-1901-05287,DBLP:journals/corr/abs-1909-02597, blimp}; (2) labeling or creating acceptability judgment resources \citep{DBLP:journals/tacl/LinzenDG16, DBLP:journals/corr/abs-1901-03438, DBLP:journals/corr/abs-1909-02597}; and (3) simulating noises for a specific NLP task such as neural machine translation \citep{DBLP:journals/corr/abs-1808-06267, anastasopoulos2019analysis}, sentiment classification \citep{DBLP:conf/eacl/BaldwinCL17}. These studies either focus on specific phenomena and mainly conduct experiments on designated corpora or rely heavily on human annotations and expert knowledge in linguistics. In contrast, our work automatically simulates natural occurring data and various types of grammatical errors and systematically analyzes how these noises affect downstream applications. This holds more practical significance to understand the robustness of several language encoders against grammatical errors.

Specifically, we first propose an effective approach to simulating diverse grammatical errors, which applies black-box adversarial attack algorithms based on real errors observed on NUS Corpus of Learner English (NUCLE) \citep{DBLP:conf/bea/DahlmeierNW13}, a grammatical error correction benchmark. This approach transforms clean corpora into corrupted ones and facilitates debugging language encoders on downstream tasks. We demonstrate its flexibility by evaluating models on four language understanding tasks and a sequence tagging task.

We next quantify model's capacities of identifying grammatical errors by probing individual layers of pre-trained encoders through a linguistic acceptability task. We construct separate datasets for eight error types. Then, we freeze encoder layers and add a simple classifier on top of each layer to predict the correctness of input texts and locate error positions. This probing task assumes if a simple classifier behaves well on a designated type of error, then the encoder layer is likely to contain knowledge of that error \citep{DBLP:conf/emnlp/ConneauKSBB17, DBLP:conf/iclr/AdiKBLG17}.

Finally, we investigate how models capture the interaction between grammatical errors and contexts. We use BERT as an example and design an unsupervised cloze test to evaluate its intrinsic functionality as a masked language model (MLM).

Our contributions are summarized as follows:
\begin{compactenum}
    \item We propose a novel approach to simulating various grammatical errors. The proposed method is flexible and can be used to verify the robustness of language encoders against grammatical errors.
    \item We conduct a systematic analysis of the robustness of language encoders and enhance previous work by studying the performance of models on downstream tasks with various grammatical error types.
    \item We demonstrate: (1) the robustness of existing language encoders against grammatical errors varies; (2) the contextual layers of language encoders acquire stronger abilities in identifying and locating grammatical errors than token embedding layers; and (3) BERT captures the interaction between errors and specific tokens in context, in particular the neighboring tokens of errors.
\end{compactenum}

The code to reproduce our experiments are available at: \url{https://github.com/uclanlp/ProbeGrammarRobustness}

\section{Related Work}
\paragraph{Probing Pre-trained Language Encoders}
The recent success of pre-trained language encoders across a diverse set of downstream tasks has stimulated significant interest in understanding their advantages. A portion of past work on analyzing pre-trained encoders is mainly based on clean data. As mentioned in \citet{DBLP:conf/acl/TenneyDP19}, these studies can be roughly divided into two categories: (1) designing controlled tasks to probe whether a specific linguistic phenomenon is captured by models \citep{DBLP:conf/acl/BaroniBLKC18, DBLP:conf/rep4nlp/PetersRS19, tenney2018what,liu-gardner-belinkov-peters-smith:2019:NAACL,
DBLP:conf/starsem/KimPPXWMTRLDBP19}, or (2) decomposing the model structure and exploring what linguistic property is encoded \citep{DBLP:conf/acl/TenneyDP19,DBLP:conf/acl/JawaharSS19, clark2019what}. However, these studies do not analyze how grammatical errors affect model behaviors.

Our work is related to studies on analyzing models with manually created noise. For example, \citet{DBLP:journals/tacl/LinzenDG16} evaluate whether LSTMs capture the hierarchical structure of language by using verbal inflection to violate subject-verb agreement. \citet{DBLP:conf/emnlp/MarvinL18} present a new dataset consisting of minimal edited pairs with the opposite linguistic acceptability on three specific linguistic phenomena and use it to evaluate RNN's syntactic ability. \citet{DBLP:journals/corr/abs-1901-05287} adjusts previous method to evaluate BERT. \citet{DBLP:journals/corr/abs-1909-02597} further compare five analysis methods under a single phenomenon. Despite the diversity in methodology, these studies share common limitations. First, they employ only a single or specific aspects of linguistic knowledge; second, their experiments are mainly based on constructed datasets instead of real-world downstream applications. In contrast, we propose a method to cover a broader range of grammatical errors and evaluate on downstream tasks. A concurrent work \citep{blimp} facilitates diagnosing language models by creating linguistic minimal pairs datasets for 67 isolate grammatical paradigms in English using linguist-crafted templates. In contrast, we do not rely heavily on artificial vocabulary and templates.

\paragraph{Synthesized Errors}
To evaluate and promote the robustness of neural models against noise, some studies manually create new datasets with specific linguistic phenomena \citep{DBLP:journals/tacl/LinzenDG16, DBLP:conf/emnlp/MarvinL18, DBLP:journals/corr/abs-1901-05287, DBLP:journals/corr/abs-1909-02597}. Others have introduced various methods to generate synthetic errors on clean downstream datasets, in particular, machine translation corpora. \citet{DBLP:conf/iclr/BelinkovB18, anastasopoulos2019analysis} demonstrate that synthetic grammatical errors induced by character manipulation and word substitution can degrade the performance of NMT systems. \citet{DBLP:conf/eacl/BaldwinCL17} augment original sentiment classification datasets with syntactically (reordering) and semantically (word substitution) noisy sentences and achieve higher performance. Our method is partly inspired by \citet{DBLP:journals/corr/abs-1808-06267}, who synthesize semi-natural ungrammatical sentences by maintaining confusion matrices for five simple error types.

Another line of studies uses black-box adversarial attack methods to create adversarial examples for debugging NLP models \citep{inproceedings-debugging,is-bert-robust, DBLP:conf/emnlp/AlzantotSEHSC18, DBLP:conf/naacl/2019-1}. These methods create a more challenging scenario for target models compared to the above data generation procedure. Our proposed simulation benefits from both adversarial attack algorithms and semi-natural grammatical errors.

\section{Method}
We first explain how we simulate ungrammatical scenarios. Then, we describe target models and the evaluation design.
\subsection{Grammatical Error Simulation}
\label{3.1}
Most downstream datasets contain only clean and grammatical sentences. Despite that recent language encoders achieve promising performance, it is unclear if they perform equally well on text data with grammatical errors.

Therefore, we synthesize grammatical errors on clean corpora to test the robustness of language encoders. We use a controllable rule-based method to collect and mimic errors observed on NUCLE following previous work \citep{DBLP:journals/corr/abs-1808-06267, sperber2017toward} and apply two ways to introduce errors to clean corpora: (1) we sample errors based on the frequency distribution of NUCLE and introduce them to plausible positions; (2) inspired by the literature of adversarial attacks \citep{inproceedings-debugging,is-bert-robust, DBLP:conf/emnlp/AlzantotSEHSC18}, we conduct search algorithms to introduce grammatical errors that causing the largest performance drop on a given downstream task.

\paragraph{Mimic Error Distribution on NUCLE} We first describe how to extract the error distribution on NUCLE \citep{DBLP:conf/bea/DahlmeierNW13}. NUCLE is constructed with naturally occurring data (student essays at NUS) annotated with error tags. Each ungrammatical sentence is paired with its correction that differs only in local edits. The two sentences make up a \emph{minimal edited pair}. An example is like:
\begin{compactenum}
    \item Will the child blame the parents after he \textbf{growing} up?$\quad \times$
    \item Will the child blame the parents after he \textbf{grows} up?$\quad \checkmark$
\end{compactenum} 
NUCLE corpus contains around 59,800 sentences with average length 20.38. About 6\% of tokens in each sentence contain grammatical errors. There are 27 error tags, including \texttt{Prep} (indicating preposition errors), \texttt{ArtOrDet} (indicating article or determiner errors), \texttt{Vform} (indicating incorrect verb form) and so forth.
 
\begin{table*}[h!]
\begin{center}
\begin{tabular}{m{2cm}m{4.6cm}m{8.0cm}}
\toprule \bf Error type & \bf Error Description &\bf Confusion Set  \\
\toprule
\texttt{ArtOrDet} & Article/determiner errors &\{ a, an, the, $\o$\} \\
\hline
\texttt{Prep} & Preposition errors &\{ on, in, at, from, for, under, over, with, into, during, until, against, among, throughout, to, by, about, like, before, across, behind, but, out, up, after, since, down, off, of, $\o$\} \\
\hline
\texttt{Trans} & Link words/phrase errors &\{and, but, so, however, as, that, thus, also, because, therefore, if, although, which, where, moreover, besides, of, $\o$\} \\
\hline
\texttt{Nn} & Noun number errors &\{SG, PL\} \\
\hline
\texttt{SVA} & Subject-verb agreement errors &\{3SG, not 3SG\} \\
\hline
\texttt{Vform} & Verb form errors &\{Present, Past, Progressive, Perfect\} \\
\hline
\texttt{Wchoice} & Word choice errors &\{Ten synonyms from WordNet Synsets\} \\
\hline
\texttt{Worder} & Word positions errors &\{Adverb w/ Adjective, Participle, Modal\} \\
\bottomrule\hline
\end{tabular}
\end{center}
\caption{The target error types and the corresponding confusion sets.}
\label{conf matrix}
\end{table*}
We consider eight frequently-occurred, token-level error types in NUCLE as shown in Table \ref{conf matrix}.

These error types perturb a sentence in terms of syntax (\texttt{SVA}, \texttt{Worder}), semantics (\texttt{Nn}, \texttt{Wchoice}, \texttt{Trans}) and both (\texttt{ArtOrDet}, \texttt{Prep}, \texttt{Vform}), and thus cover a wide range of noise in natural language. Then, we construct a confusion set for each error type based on the observation on NUCLE. Each member of a confusion set is a token. We assign a weight $w_{ij}$ between token $t_i$ and $t_j$ in the same set to indicate the probability that $t_i$ will be replaced by $t_j$. In particular, for \texttt{ArtOrDet}, \texttt{Prep} and \texttt{Trans}, the confusion set consists of a set of tokens that frequently occur as errors or corrections on NUCLE. For each token $t_i$ in the set, we compute $w_{ij}$ based on how many times $t_i$ is replaced by $t_j$ in minimal edited pairs on NUCLE.

Notice that we add a special token $\o$ to represent deletion and insertion. For \texttt{Nn}, when we find a noun, we add it and its singular (SG) or plural (PL) counterpart to the set. For \texttt{SVA}, when we find a verb with present tense, we add it and its third-person-singular (3SG) or non-third (not 3SG) counterpart to the set. For \texttt{Worder}, we exchange the position of an adverb with its neighboring adjective, participle or modal. For \texttt{Vform}, we use NLTK \citep{nltk} to extract present, past, progressive, and perfect tense of a verb and add to the set. For \texttt{Wchoice}, we select ten synonyms of a target word from WordNet. The substitution weight is set to be uniform for both \texttt{Vform} and \texttt{Wchoice}.

\paragraph{Grammatical Error Introduction}  We introduce errors in two ways. The first is called \emph{probabilistic transformation}. Similar to \citet{DBLP:journals/corr/abs-1808-06267}, we first obtain the parse tree of the target sentence using the Berkeley syntactic parser \citep{DBLP:conf/acl/PetrovBTK06}. Then, we sample an error type from the error type distribution estimated from NUCLE and randomly choose a position that can apply this type of error according to the parse tree. Finally, we sample an error token based on the weights from the confusion set of the sampled error type and introduce the error token to the selected position. 

However, \emph{probabilistic transformation} only represents the average case. To debug and analyze the robustness of language encoders, we consider another more challenging setting -- \emph{worst-case transformation}, where we leverage search algorithms from the black-box adversarial attack to determine error positions. More concretely, we obtain an operation set for each token in a sentence by considering all possible substitutions based on all confusion sets. Note that some confusion sets are not applicable, for example the confusion set of \texttt{Nn} to a verb. Each operation in the operation set is to replace the target token or to change its position. Then, we apply a searching algorithm to select operations from these operation sets that change the prediction of the tested model and apply them to generate error sentences. Three search algorithms are considered: \emph{greedy search}, \emph{beam search}, and \emph{genetic algorithm}. 

\emph{Greedy search} attack is a two-step procedure. First, we evaluate the importance of tokens in a sentence. The importance of a token is represented by the likelihood decrease on the model prediction when it is deleted. The larger the decrease is, the more important the token is. After comparing all tokens, we obtain a sorted list of tokens in descending order of their importance. Then, we walk through the list. For each token in the list, we try out all operations from the operation set associated with that token and then practice the operation that degrades the likelihood of the model prediction the most. We keep repeating step two until the prediction changes or a budget (e.g., number of operations per sentence) is reached. 

\emph{Beam search} is similar to \emph{greedy search}. The only difference is that when we walk through the sorted list of tokens, we maintain a beam with fixed size $k$ that contains the top $k$ operation streams with the highest global degradation. 

\emph{Genetic algorithm} is a population-based iterative method for finding more suitable examples. We start by randomly selecting operations to build a generation and then use a combination of crossover and mutation to find better candidates. We refer the readers to \citet{DBLP:conf/emnlp/AlzantotSEHSC18} for details of the genetic algorithm in adversarial attack. Comprehensive descriptions of all methods are found in Appendix \ref{Attack}.

\subsection{Target Models}
\label{3.2}
We evaluate the following three pre-trained language encoders. Detailed descriptions of models and training settings are in Appendix \ref{model}.

\paragraph{ELMo}\citep{DBLP:conf/naacl/PetersNIGCLZ18}\quad is a three-layer LSTM-based model pre-trained on the bidirectional language modeling task on 1B Word Benchmark \citep{DBLP:conf/interspeech/ChelbaMSGBKR14}. We fix ELMo as a contextual embedding and add two layers of BiLSTM with attention mechanism on top of it.
\paragraph{BERT}\citep{DBLP:conf/naacl/DevlinCLT19}\quad is a transformer-based \citep{DBLP:conf/nips/VaswaniSPUJGKP17} model pre-trained on masked language modeling and next sentence prediction tasks. It uses 16GB English text and adapts to downstream tasks by fine-tuning. We use \textit{BERT-base-cased} for Named Entity Recognition (NER) and \textit{BERT-base-uncased} for other tasks and perform task-specific fine-tuning. 
\paragraph{RoBERTa} \citep{DBLP:journals/corr/abs-1907-11692}\quad is a robustly pre-trained BERT model using larger pre-training data (160GB in total), longer pre-training time, the dynamic masking strategy and other optimized pre-training methods. We use \textit{RoBERTa-base} and perform task-specific fine-tuning.

\subsection{Evaluation Methods}
\label{3.3}
We design the following three evaluation methods to systematically analyze how language encoders are affected by grammatical errors in input.

\paragraph{Simulate Errors on Downstream Tasks} Using the simulation methods discussed in Section \S\ref{3.1}, we are able to perform evaluation on existing benchmark corpora. In our experiments, we consider the target models independently. The whole procedure is: given a dataset, the target model is first trained (fine-tuned) and evaluated on the clean training and development set. Then, we discard those wrongly predicted examples from the development set and apply simulation methods to perturb each remaining example. We compute the attack success rate (attacked examples / all examples) as an indicator of model robustness against grammatical errors. The smaller the rate is, the more robust a model is.

\paragraph{Linguistic Acceptability Probing} We design a linguistic acceptability probing task to evaluate each individual type of error. We consider two aspects: (1) if the model can tell whether a sentence is grammatically correct or not (i.e., a binary classification task); (2) if the model can locate error positions in the token-level. We fix the target model and train a self-attention classifier to perform both probing tasks.

\paragraph{Cloze test for BERT} We design an unsupervised cloze test to evaluate the masked language model component of BERT based on minimal edited pairs. For each minimal pair that differs only in one token, we quantify how the probability of predicting a single masked token in the rest of the sentence affected by this grammatical error. This method analyzes how error token affects clean context, which is complementary to \citet{DBLP:journals/corr/abs-1901-05287} who focuses on \texttt{SVA} error and discusses how clean contexts influence the prediction of the masked error token.

\begin{table}[t!]
\begin{center}
\small
\begin{tabular}{m{1.5cm}<{\centering}m{1.0cm}<{\centering}m{0.9cm}<{\centering}m{0.9cm}<{\centering}m{1.2cm}<{\centering}}
\toprule & \bf InferSent & \bf ELMo & \bf BERT & \bf RoBERTa\\ \toprule
\bf MRPC & 75.42 & 80.30 & 86.48 & 89.88\\
\bf MNLI-m & 68.62 & 74.91 & 83.77 & 87.70\\
\bf MNLI-mm & 69.12 & 75.50 & 84.80 & 87.40\\
\bf QNLI & 77.39 & 78.23 & 90.58 & 92.50\\
\bf SST-2 & 83.14 & 90.37 & 92.08 & 94.72\\
\bf NER & - & 91.21 & 95.20 & 95.45 \\
\bottomrule\hline
\end{tabular}
\end{center}
\caption{\label{font-table} Original performance of the target models on language understanding and sequential tagging tasks.}
\label{origin_results}
\end{table}
\begin{table*}[h!]
\centering
\small
\begin{tabular}{m{1.05cm}m{1.35cm}<{\centering}m{1.7cm}<{\centering}m{3.9cm}<{\centering}m{1.7cm}<{\centering}m{1.7cm}<{\centering}m{1.7cm}<{\centering}}
  \toprule
  \bf{Model} & \bf{Alg.}  & \bf{MRPC} & \bf{MNLI (m/mm)} & \bf{QNLI} & \bf{SST-2} & \bf{NER}\\
  \toprule
  InferSent & dist. & 6.51 (14.53) & 8.30 (13.98) / 8.80 (14.23) & 4.76 (12.53) & 5.79 (14.38) & - \cr
  & greedy & 53.42 (9.02) &  36.52 (10.35) / 40.71 (10.06) & 44.92 (7.61) & 43.44 (8.02) & -\cr
  & beam & 54.39 (9.08)& 36.66 (10.37) / 40.87 (10.06) & 45.16 (7.62) & 43.86 (8.03) & -\cr
  & genetic & 79.15 (8.60) & - & - & 59.86 (8.39) & -\cr
  \midrule
  BiLSTM & dist. & 9.99 (14.53) & 7.76 (13.98) / 7.83 (14.23) & 5.34 (12.53) & 4.64 (14.38) & 3.29 (13.75) \cr
  + ELMo & greedy & 60.84 (8.19) & 29.58 (10.28) / 32.92 (9.89) & 39.12 (7.25) & 37.55 (8.24) & 17.81 (7.67)\cr
  + Attn & beam & 61.49 (8.29) & 29.74 (10.29) / 33.12 (9.91) & 40.38 (7.33) & 38.32 (8.32) & 18.33 (7.85) \cr
  & genetic & 81.14 (7.41) & - & - & 59.25 (8.25) & 39.78 (8.19)\cr
  \midrule
  BERT & dist. & 3.69(14.53) & 6.59 (13.98) / 6.95 (14.23) & 2.33 (12.53) & 4.73 (14.38) & 3.07 (13.75) \cr
  & greedy & 31.25 (7.95)&  28.76 (10.28) / 32.04 (10.01)  & 25.43 (7.38) & 33.54 (7.96) & 17.12 (7.51) \cr
  & beam & 31.81 (8.01) & 29.03 (10.30) / 32.44 (10.04) & 26.42 (7.48) & 34.28 (8.01) & 18.27 (7.74) \cr
  & genetic & 59.01 (8.84) & - & - & 58.53 (7.83) & 38.83(7.64) \cr
  \midrule
  RoBERTa & dist. & 3.04 (14.53) & 5.66 (13.98) / 5.88(14.23) & 1.92 (12.53) & 3.53 (14.38) & 2.52 (13.75) \cr
  & greedy & 20.45 (8.11) & 20.65 (10.43) / 21.47 (10.02) & 19.82 (7.18) & 31.06 (8.20) & 15.84 (8.12)\cr
  & beam & 20.73(8.14) & 20.89 (10.44) / 21.91 (10.06) & 20.52 (7.29) & 31.91 (8.27) & 16.51 (7.47) \cr
  & genetic & 38.93 (9.17) & - & - & 56.41 (8.39) & 35.11(7.55) \cr
  \bottomrule\hline

\end{tabular}
\caption{Results of evaluating the robustness of models on downstream tasks. Each column represents a dataset and each row represents a victim model with the attack algorithm (dist. means \emph{probabilistic transformation}). In each cell, we show the mean attack success rate (in percentage) and the mean percentage of modified words (number in the bracket) over the dataset.}
\label{attacked results GLUE}
\end{table*}

\section{How Grammatical Errors Affect Downstream Performance?}
In this section, we simulate grammatical errors and analyze performance drops on downstream tasks.

We compare ELMo, BERT, RoBERTa and a baseline model InferSent \citep{DBLP:conf/emnlp/ConneauKSBB17}.
\paragraph{Datasets} We use four language understanding datasets: MRPC \citep{DBLP:conf/acl-iwp/DolanB05}, MNLI \citep{DBLP:conf/naacl/WilliamsNB18}, QNLI \citep{DBLP:conf/emnlp/RajpurkarZLL16}, and SST-2 \citep{DBLP:conf/emnlp/SocherPWCMNP13} from GLUE \citep{DBLP:conf/iclr/WangSMHLB19} and a sequence tagging benchmark: CoNLL-2013 for NER. Detailed descriptions of these corpora are in Appendix \ref{Dataset}. We do not use other datasets from GLUE since they are either small in size or only contain short sentences.
\paragraph{Attack Settings} For all tasks, we limit the maximum percentage
of allowed modifications in a sentence to be 15\% of tokens, which is a reasonable rate according to the statistics estimated from the real data. As shown in Table \ref{attacked results GLUE}, the \emph{worst-case transformation} only modifies around 9\% of tokens overall under such a limitation. For MNLI and QNLI, we only modify the second sentence, i.e., hypothesis and answer, respectively. For MRPC, we only modify the first sentence. We do not apply the genetic algorithm to MNLI and QNLI due to their relatively large number of examples in the development sets, which induce an extremely long time for attacking. For NER, we keep the named entities and only modify the remaining tokens. The data evaluation and some examples are shown in Appendix \ref{example}

\paragraph{Results and Discussion}
Table \ref{origin_results} presents the test performance of four target models on the standard development set of each task. Table \ref{attacked results GLUE} summarizes the attack success rates on language understanding tasks, the decreases of F1 score on NER, and the mean percentage of modified tokens (number in brackets). All numbers are formatted in percentage.

As shown in Table \ref{attacked results GLUE}, with the \emph{probabilistic transformation}, the attack success rates fall between 2\% (RoBERTa, QNLI) and 10\% (ELMo, MRPC). With the \emph{worst-case transformation}, we obtain the highest attacked rate of 81.1\% (ELMo, genetic algorithm, MRPC) and an average attacked rate across all tasks of 29\% by perturbing only around 9\% of tokens. This result confirms that all models are influenced by ungrammatical inputs. NER task is in general harder to be influenced by grammatical errors. In terms of the \emph{probabilistic transformation}, the drop of F1 scores ranges from 2\% to 4\%. For the \emph{worst-case transformation}, the highest drop for NER is 18.33\% (ElMo, beam search).

\begin{table}[t]
\begin{center}
\small
\begin{tabular}{m{1.0cm}m{0.6cm}<{\centering}m{0.7cm}<{\centering}m{0.5cm}<{\centering}m{0.6cm}<{\centering}m{0.7cm}<{\centering}m{0.5cm}<{\centering}}
\toprule
& \multicolumn{3}{c}{BERT} & \multicolumn{3}{c}{RoBERTa}
\\
\cmidrule{2-4}\cmidrule{5-7}
& \bf MRPC & \bf MNLI & \bf SST & \bf MRPC & \bf MNLI & \bf SST \\ 
\toprule
\bf Prep &16&178&36&15&103&43\\
\bf Art/Det & 5 & 270 & 20 & 7 & 228 & 28\\
\bf Wchoice & 93 & 1129 & 233 & 64 & 772 & 195\\
\bf Vform & 8 & 231 & 26 & 9 & 314 & 37\\
\bf SVA & 57 & 538 & 83 & 31 & 388 & 83\\
\bf Nn & 14 & 128 &13 & 3 & 84 &13\\
\bf Worder & 0 & 62 &28 & 0 & 43 &28\\
\bf Trans & 5 & 70 & 25 & 5 & 31 & 25\\
\bottomrule\hline
\end{tabular}
\end{center}
\caption{\label{font-table} Numbers of times each error type is chosen in successful attacks. We find that \texttt{Wchoice} and \texttt{SVA} are more harmful.}
\label{error type identify}
\end{table}
Considering different target models, we observe that the impact of grammatical errors varies among models. Specifically, RoBERTa exhibits a strong robustness against the impact of grammatical errors, with consistently lower attack success rates (20.28\% on average) and F1 score decreases (17.50\% on average) across all tasks, especially on MRPC and MNLI. On the other hand, BERT, ELMo, and InferSent experience an average attack rate of 26.03\%, 33.06\%, 36.07\% respectively on NLU tasks. Given the differences in pre-training strategies, we speculate that pre-training with more data might benefit model robustness against noised data. This speculation is consistent with \citep{blimp}, where the authors also give a lightweight demonstration on LSTM and Transformer-XL \citep{transformer-XL} with varying training data. We leave a further exploration of this speculation and a detailed analysis of model architecture to future work.

Note that in the experiment setting, for each model, we follow the literature to compute the attack success rate only on the instances where the model makes correct predictions. Therefore, the attack success rates across different models are not comparable. To compare the robustness of different encoders, we further examine the attack success rates on the common part in the development set that all the models make correct predictions. We find that the overall trend is similar to that in Table \ref{attacked results GLUE}. For example, the greedy attack success rates of RoBERTa, BERT, and ELMo on MRPC and SST-2 are 14.4\%, 22.1\%, 46.8\%, and  28.2\%, 30.0\%, 33.9\% respectively.

To better understand the effect of grammatical errors, we also analyze (1) which error type harms the performance most, (2) how different error rates affect the performance. For the first question, we represent the harm of an error type by the total time it is chosen in successful greedy attack examples. We conduct experiments to analyze BERT and RoBERTa on the development sets of MRPC, MNLI-m, and SST-2 as shown in Table \ref{error type identify}. Among all, \texttt{Wchoice} is the most harmful type while \texttt{Worder} the least. \texttt{SVA} ranks the second most harmful type. Notice that though \texttt{Nn} changes a token in a similar way with \texttt{SVA} (both adding or dropping -s or -es in most cases), they have different influences to the model. As for errors related to function words, \texttt{Prep} plays a more important role in general but \texttt{ArtOrDet} harms MNLI more.

\begin{figure}[t]
\centering
\includegraphics[scale=0.40]{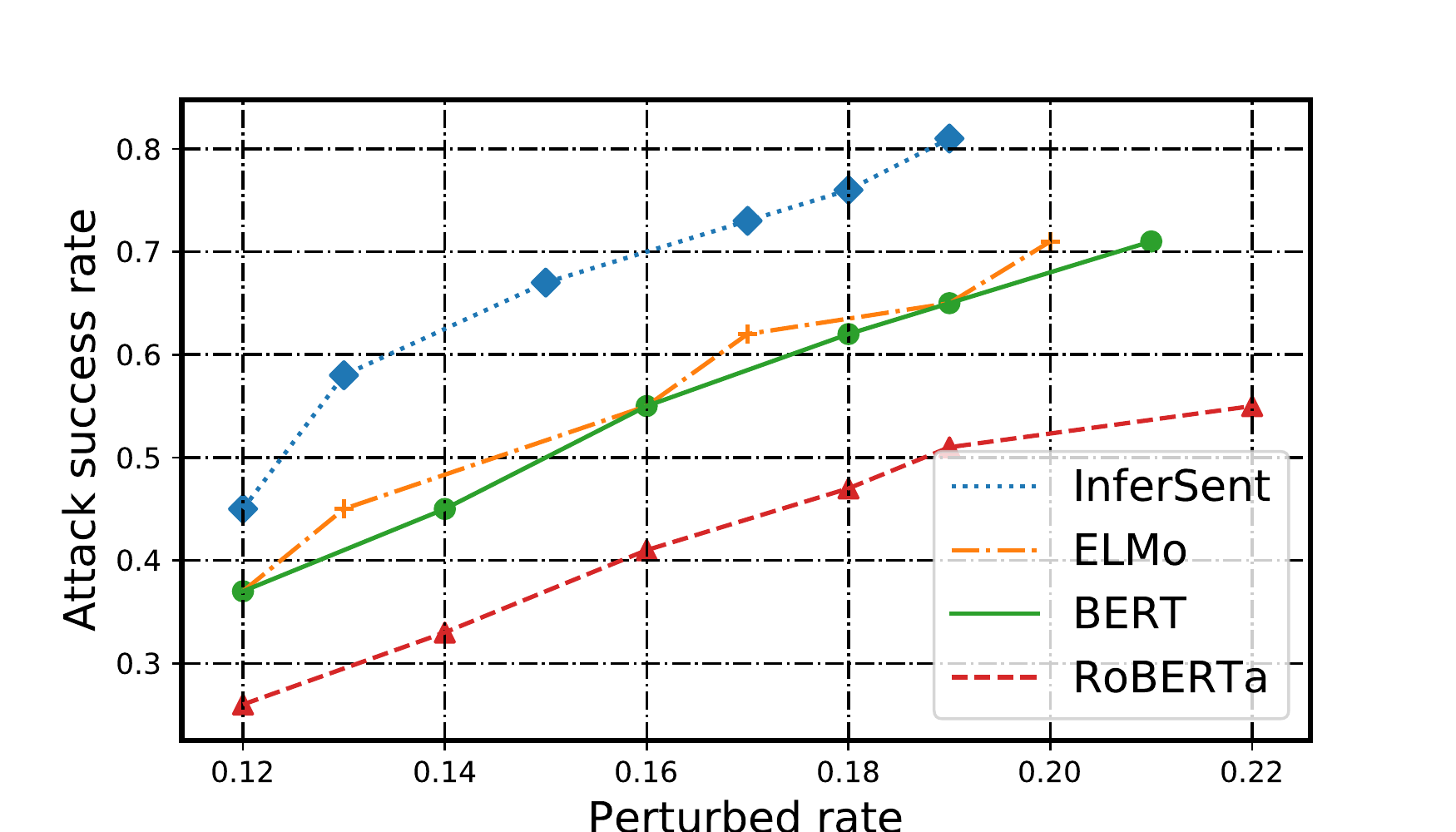}
\caption{Attack success rate when the numbers of modified tokens in a sentence increase. }
\label{vis_rate}
\end{figure}

For the second one, we increase the allowed modifications of greedy attack from 15\% to 45\% of tokens in one sentence, resulting the actual percentage of modified tokens under 20\%. We evaluate all models on the development set of MNLI-m. Results are shown in Fig \ref{vis_rate}. We find that all attack success rates grow almost linearly as we increase modifications. ELMo and BERT perform almost the same while InferSent grows faster at the beginning and RoBERTa grows slower when reaching the end. The average attack success rate comes to 70\% when the error rate is around 20\%.

\section{To What Extent Models Identify Grammatical Errors?}

Our goal in this section is to assess the ability of the pre-trained encoders in identifying grammatical errors. We use a binary linguistic acceptability task to test the model ability in judging the grammatical correctness of a sentence. We further study whether the model can precisely locate error positions, which reflects the token-level ability.
\paragraph{Data} We construct separate datasets for each specific type of grammatical error. For each dataset, we extract 10,000 sentences whose lengths fall within 10 to 60 tokens from 1B Word Benchmark \citep{DBLP:conf/interspeech/ChelbaMSGBKR14}. Then, we introduce the target error type to half of these sentences using \emph{probabilistic transformation} and keep the error rate over each dataset around 3\% (resulting in one or two errors in each sentence). Sentences are split into training (80\%), development (10\%) and test (10\%).
\paragraph{Models} We study individual layers of ELMo (2 layers), BERT-base-uncased (12 layers) and RoBERTa-base (12 layers). In particular, we fix each layer and attach a trainable self-attention layer on top of it to obtain a sentence representation. The sentence representation is fed into a linear classifier to output the probability of whether the sentence is linguistically acceptable. See details about the self-attention layer and the linear classifier in Appendix \ref{selfattnmodel}. We next extract the top two positions with the heaviest weights from the trained self-attention layer. If the positions with error token are included, we consider the errors are correctly located by the model in the token-level. This suggests whether contextual encoders are providing enough information for the classifier to identify error locations. For comparisons, we also evaluate the input embedding layer (non-contextualized, layer 0) of each model as a baseline. We compute accuracy for both sentence-level and token-level evaluations.

\begin{figure}[t]
\includegraphics[width=78 mm, height = 48mm]{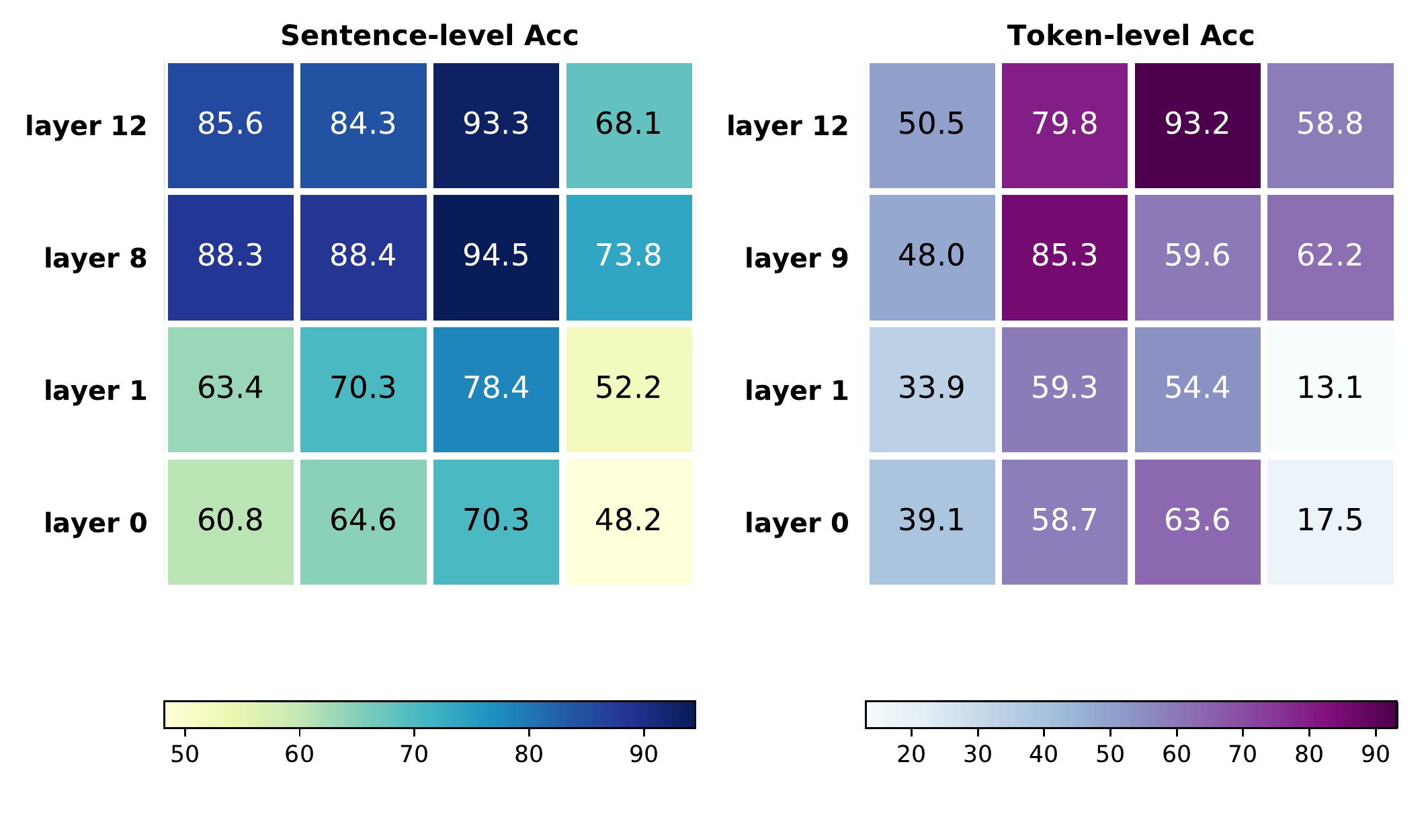}
\caption{Probing four layers of BERT on four error types. The left side shows the accuracy of the binary linguistic acceptability task. The right side shows the accuracy of locating error positions. Each row represents a specific layer, and each column represents a type of errors, \texttt{ArtOrDet}, \texttt{Nn}, \texttt{SVA}, \texttt{Worder} from left to right. Full results are given in Appendix \ref{probe}}
\label{visual layers}
\end{figure}

\paragraph{Results and Discussion}
We visualize the results of four layers of BERT on four error types, \texttt{ArtOrDet}, \texttt{Nn}, \texttt{SVA}, and \texttt{Worder} in Fig \ref{visual layers}. Complete results of all layers and other error types are in Appendix \ref{probe}. We find that the mean sentence-level accuracy of the best contextual layers of BERT, ELMo, and RoBERTa across error types are 87.8\%, 84.3\%, and 90.4\%, respectively, while input embedding layers achieve 64.7\%, 65.8\%, and 66.0\%. In token-level, despite trained only on the prediction of whether a sentence is acceptable, the mean accuracy of classifiers upon the best layers of BERT, ELMo, and RoBERTa are 79.3\%, 63.3\%, and 80.3\%, compared to 48.6\%, 18.7\%, and 53.4\% of input embedding layers. The two facts indicate that these pre-trained encoder layers possess stronger grammatical error detecting and locating abilities compared to input embedding layers.

\begin{figure}[t]
\centering\includegraphics[width=1.07\linewidth]{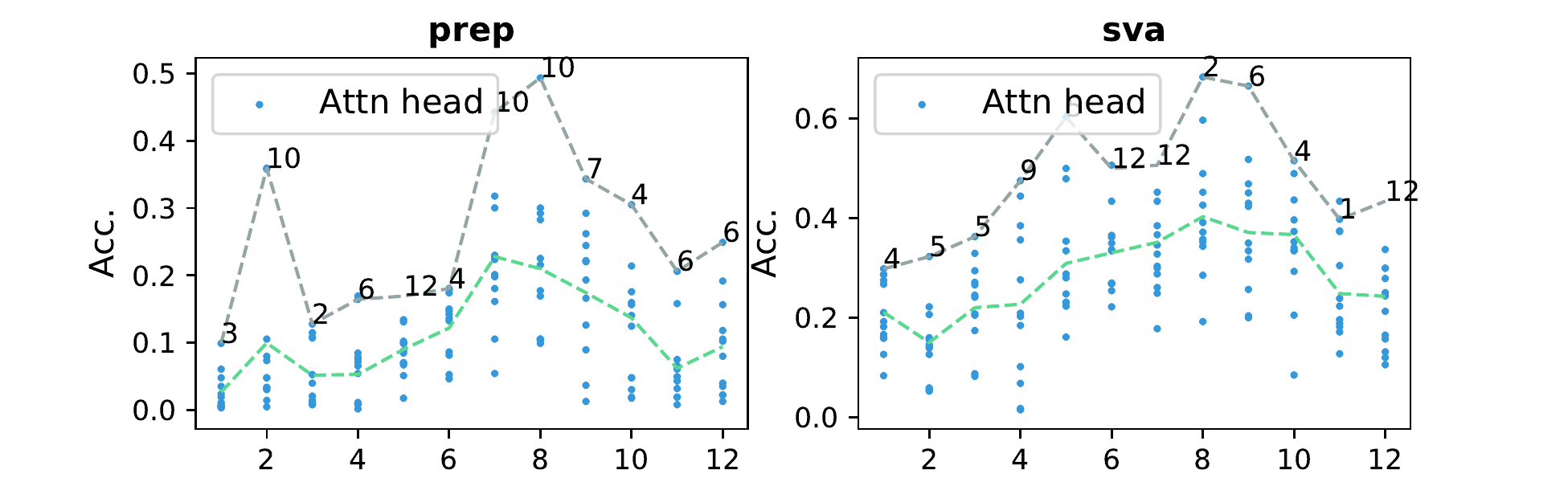}
\caption{The accuracy of each attention head of BERT on token-level evaluation. The grey line stands for the best performing heads. The green line stands for the average performance of heads in one layer.}
\label{pahead}
\end{figure}

We also observe patterns related to a specific model. Specifically, middle layers (layers 7-9) of BERT are better at identifying errors than lower or higher layers, as shown in Fig \ref{visual layers}. But higher layers of BERT locate errors related to long-range dependencies and verbs such as \texttt{SVA} and \texttt{Vform} more accurately. To further investigate BERT's knowledge of error locations. We conduct the same token-level evaluation to the 144 attention heads in BERT. Results for \texttt{Prep} and \texttt{SVA} are visualized in Fig \ref{pahead}. We find that even in a completely unsupervised manner, some attention heads results for 50\%-60\% accuracy in locating errors. Consistent with self-attention layers, attention heads from middle layers perform the best. See Appendix \ref{heads} for all error types.

Due to space limit, we present results of RoBERTa and ELMo in Appendix \ref{probe} and summarize the observations in the following. RoBERTa exhibits better ability in detecting and locating errors in lower layers compared to BERT and achieves the best performance in top layers (layers 10-11). It is also good at capturing verb and dependency errors. On the other hand, the first layer of ELMo consistently gives the highest sentence-level classification accuracy. But its best performing layer in locating errors depends on the error type and varies between the first and the second layer. In particular, The second layer of ELMo exhibits strong ability in locating \texttt{Nn} and outperforms BERT in accuracy. This is surprising given the fact that \texttt{Nn} is not obvious with character embeddings from layer 0 of ELMo. We further notice that for all models, \texttt{Worder} is the hardest type to detect in the sentence-level and \texttt{ArtOrDet} and \texttt{Worder} are the hardest types to locate in the token-level. We hypothesize this is related to the locality of these errors which induces a weak signal for models to identify them.
Appendix \ref{case study} demonstrates some examples of the token-level evaluation of BERT.

\section{How BERT Captures the Interaction between Tokens When Errors Present}
We aim to reveal the interaction between grammatical errors and their nearby tokens through studying the masked language model (MLM) component of BERT. We investigate BERT as it is a typical transformer-based encoder. Our analysis can be extended to other models.  
\paragraph{Experimental Settings} We conduct experiments on minimal edited pairs from NUCLE. We extract pairs with error tags \texttt{ArtOrDet}, \texttt{Prep}, \texttt{Vt}, \texttt{Vform}, \texttt{SVA}, \texttt{Nn}, \texttt{Wchoice}, \texttt{Trans} and keep those that only have one token changed. This gives us eight collections of minimal edited pairs with sizes of 586, 1525, 1817, 943, 2513, 1359, 3340, and 452, respectively.

Given a minimal edited pair, we consider tokens within six-token away from the error token. We replace the same token in the grammatical and ungrammatical sentence with \texttt{[MASK]} one at a time and use BERT as an MLM to predict its likelihood. Then we compute the likelihood drop in the ungrammatical sentence and obtain the average drop over all minimal edited pairs.
\begin{figure}[t]
\includegraphics[scale=0.37]{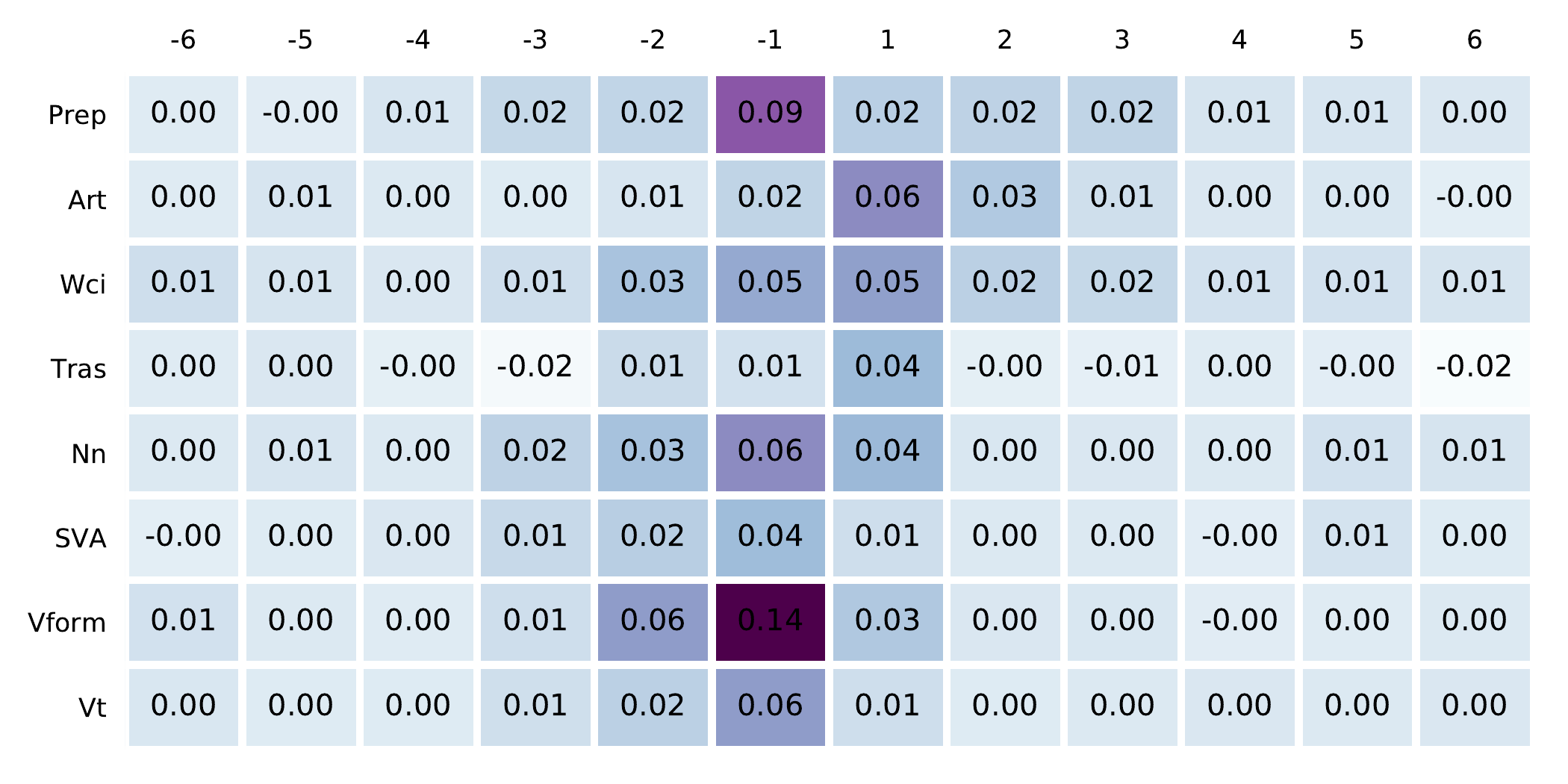}
\centering
\caption{Probing BERT as an MLM. Each row represents a target error type. Each column represents the distance from the error position. Each number represents the mean likelihood drop over all pairs. We find that specific tokens are affected more by error tokens.}
\label{visual mlm}
\end{figure}

\begin{table}[t] 
\begin{center}
\small
\begin{tabular}{m{0.3cm}m{6.5cm}}
\toprule
$\checkmark$ & This would thus reduce the financial burden of \textcolor{red}{this} \textcolor{blue}{group} of people based on their income ceilings .\cr
$\times$ & This would thus reduce the financial burden of \textcolor{red}{these} \textcolor{blue}{group} of people based on their income ceilings . \cr
\end{tabular}
\begin{tabular}{m{0.1cm}*{6}{c}}
\ldots & burden & of & \textcolor{red}{this (these)} & \textcolor{blue}{group}  & of & \ldots \cr
\ldots & 0.01 & 0.09 & - & 0.41  & 0.02 & \ldots \cr
\end{tabular}

\begin{tabular}{m{0.3cm}m{6.5cm}}
\midrule
$\checkmark$ & The inexpensive fuel cost and the sheer volume of energy produced by \textcolor{red}{a} nuclear \textcolor{blue}{reactor} far outweighs the cost of research and development . \cr
$\times$ & The inexpensive fuel cost and the sheer volume of energy produced by \textcolor{red}{the} nuclear \textcolor{blue}{reactor} far outweighs the cost of research and development . \cr
\end{tabular}
\begin{tabular}{m{0.2cm}*{6}{c}}
\ldots & produced & by & \textcolor{red}{a (the)} & nuclear & \textcolor{blue}{reactor} & \ldots \cr
\ldots & 0.05 & -0.02 & - & 0.31  & 0.42 & \ldots \cr
\bottomrule\hline
\end{tabular}

\end{center}
\caption{Examples with \texttt{ArtOrDet}. We show the minimal edit pairs and the likelihood decrease of each token within two tokens away from the errors. Wrong determiners and their corrections are marked in red. The heads in determiner-noun dependencies are marked in blue. As shown in the table, the heads in determiner-noun dependencies get the largest likelihood decrease.}
\label{example}
\end{table}

\paragraph{Results and Discussion}
Results are visualized in Fig\ \ref{visual mlm}. In general, We find that the decrease of likelihood on specific positions are greater than others in the presence of errors. Given the fact that certain dependencies between tokens such as subject-verb and determiner-noun dependencies are accurately modeled by BERT as demonstrated in prior work \citep{DBLP:conf/acl/JawaharSS19}, we suspect that the presence of an error token will mostly affect its neighboring tokens (both in terms of syntactic and physical neighbors). This is consistent with our observation in Fig\ \ref{visual mlm} that in the case of \texttt{SVA} where a subject is mostly the preceding token of a verb (although agreement attractors can exist between subject and verb), the proceeding tokens of error positions get the largest likelihood decreases overall. In the case of \texttt{ArtOrDet} where an article or a determiner can be an indicator and a dependent of the subsequent noun, predicting the next tokens of error positions becomes much harder. We provide two running examples with \texttt{ArtOrDet} in Table \ref{example} to further illustrate this point.

\section{Adversarial Training}
Finally, we explore a data augmentation method based on the proposed grammatical error simulations. We apply the greedy search algorithm to introduce grammatical errors to the training examples of a target task and retrain the model on the combination of original examples and the generated examples. We take the MRPC \citep{DBLP:conf/acl-iwp/DolanB05} dataset as an example to demonstrate the results. We augment the training set of MRPC with different proportions of adversarial examples, fine-tune BERT on the augmented training set and then evaluate on both the original development set and the corrupted development set.

\begin{figure}[t]
\centering
\includegraphics[scale=0.405]{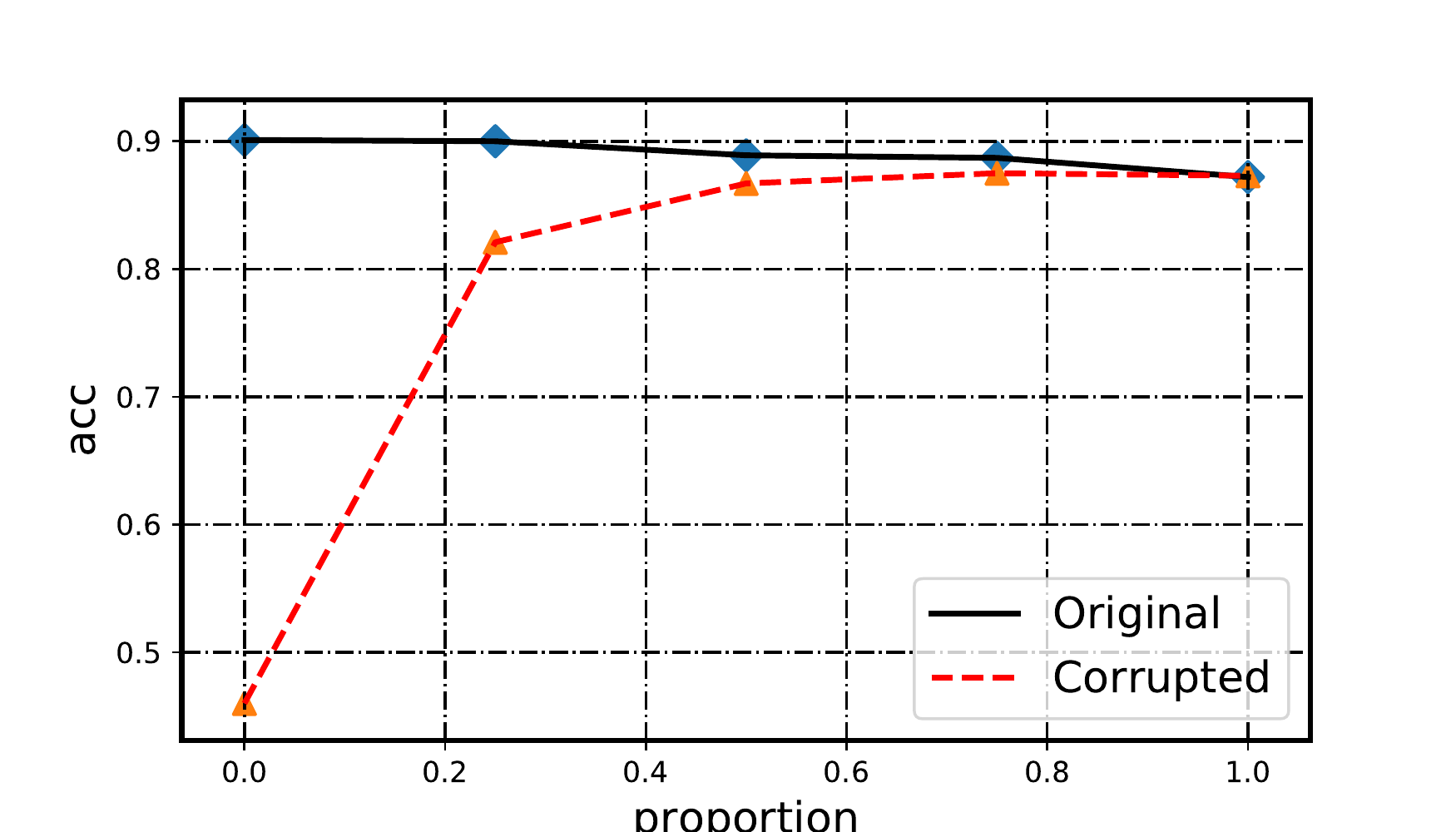}
\caption{Results of a data augmentation defense. The proportions indicate the amount of adversarial examples augmented to the training set compared to original amount. The dash and solid lines show the accuracy on corrupted and original development set with different proportions respectively.}
\label{vis_da}
\end{figure}

Results are shown in Figure \ref{vis_da}. we find that by adding a small number of adversarial examples, the accuracy is recovered from 46\% to 82\%. As the proportion of augmented adversarial examples increases, the accuracy continues to increase on the corrupted set, with negligible changes to the original validation accuracy. This fact also demonstrates that our simulated examples are potentially helpful for reducing the effect of grammatical errors. 
\section{Conclusion}
In this paper, we conducted a thorough study to evaluate the robustness of language encoders against grammatical errors. We proposed a novel method to simulating grammatical errors and facilitating our evaluations. We studied three pre-trained language encoders, ELMo, BERT, and RoBERTa and concentrated on three aspects of their abilities against grammatical errors: performance on downstream tasks when confronted with noised texts, ability in identifying errors and capturing the interaction between tokens in the presence of errors. This study shed light on understanding the behaviors of language encoders against grammatical errors and encouraged future work to enhance the robustness of these models.

\paragraph{Acknowledgements}
We would like to thank the anonymous reviewers for their feedback. This work is supported by NSF Grant \#IIS-1927554.

\bibliography{acl2020}
\bibliographystyle{acl_natbib}
\clearpage
\appendix
\section{Downstream Task Details}
\label{Dataset}
We test on four language understanding and a sequence labeling datasets. Statistics of these datasets are listed in Table \ref{datatable}.
\paragraph{MRPC} The Microsoft Research Paraphrase Corpus (MRPC) \citep{DBLP:conf/acl-iwp/DolanB05} is a paraphrase detection task which aims to predict a binary label for whether two sentences are semantically equivalent.
\paragraph{MNLI}\quad The Multi-Genre Natural Language Inference Corpus (MNLI) \citep{DBLP:conf/naacl/WilliamsNB18} is a broad-domain natural language inference task to predict the relation (entailment, contradiction, neutral) between premise and hypothesis. MNLI contains both the matched (in-domain) and mismatched (cross-domain) sections.
\paragraph{QNLI} The Question-answering NLI task (QNLI) is recasted from the Stanford Question Answering Dataset \citep{DBLP:conf/emnlp/RajpurkarZLL16}, which aims to determine whether a context sentence contains
the answer to the question (entailment, not entailment). 
\paragraph{SST-2} The Stanford Sentiment Treebank two-way class split (SST-2; \citep{DBLP:conf/emnlp/SocherPWCMNP13}) is a binary classification task which assigns positive or negative labels to movie review sentences.
\paragraph{CoNLL2003 - NER} The shared task of CoNLL-2003 Named Entity Recognition (NER) \citep{DBLP:journals/corr/cs-CL-0306050} is a token level sequence labeling task to recognize four types of named entity: persons, locations, organizations and names of miscellaneous entities that do not belong to the previous three groups.
\begin{table}[h]
\centering
\begin{tabular}{cccc}
  \toprule
  Dataset & Train & Dev & Avg Len \cr
  \toprule
  MRPC & 3.7k & 409 & 22.4\cr
  MNLI & 393k & 19k & 10.1\cr
  QNLI & 105k & 5.5k & 27.6\cr
  SST-2 & 67k & 873 & 19.5\cr
  CoNLL2003 & 15k & 3k & 14.8\cr
  \bottomrule\hline

\end{tabular}
\caption{Datasets statistics of MRPC, MNLI, QNLI, SST-2, and CoNLL2003. Train and Dev stands for the number of sentences in the train and development set. Avg Len stands for the average sentence length (in token) of the target sentence being attacked.}
\label{datatable}
\end{table}

\section{Model Details}
\label{model}
\subsection{Pre-trained Encoder Details}

We study BERT (base, uncased), BERT (base, cased) (for NER only), RoBERTa (base), and ELMo. BERT (base) and RoBERTa (base) have the same architecture. Both of them are deep transformer models with 12 layers and 12 attention heads, 768 hidden size in each layer. They contain a learnable output layer for fine-tuning on \texttt{[CLS]} or \texttt{<s>}. We use PyTorch implement of BERT and RoBERTa from \citet{Wolf2019HuggingFacesTS} and fine-tune them on downstream tasks. For ELMo, we fix ELMo representations as contextual embeddings of tokens and feed them to a two-layer, 1500D BiLSTM with cross-sentence attention mechanism as implemented in \textit{jiant}. \citep{wang2019jiant}.
\subsection{Training and Fine-tuning Details}
For BERT and RoBERTa, we set the maximum input length to be 128, the maximum number of epochs to be 3, and the dropout to be 0.1 for all tasks. We use Adam \citep{adam} with an initial learning rate of 2e-5, batch size 16 and no warm-up steps for training. For ELMo, we train the BiLSTM using Adam \citep{adam} with an initial learning rate of 1e-4, batch size 32. We set the dropout to be 0.2, the maximum number of epochs to be 10 and divide the learning rate by 5 when the performance does not improve for 2 epochs. 
\subsection{Probing model Details}
\label{selfattnmodel}
We use a self-attention layer and a linear classifier to compose the probing component in section 5. The self-attention layer takes as input the hidden representations from the fixed layer $i$ of an encoder, denoted as $h = \{h_{1}^{i}, h_{2}^{i}, ..., h_{n}^{i}\}$ and outputs a sentence representation $s_{i}$:
\begin{equation}
s_{i} = \Sigma_{j=1}^{n}\alpha_{j}h_{j}^{i}
\end{equation}
\begin{equation}
\alpha_{j} = softmax(v_b^{T}tanh(W_ah_{j}^{i})) 
\end{equation}
where $W_a$ is a weight matrix and $v_b$ is a vector of parameters. $s_i$ is fed to the classifier to output the probability of the sentence being linguistically acceptable. The self-attention layer has a hidden dim of 100 and 0.1 dropout. The classifier has 1 layer and 0.1 dropout. The probing model is trained with Adam \citep{adam} using a learning rate of 0.001, batch size of 8 , $L_2$ weight decay of 0.001 for 10 epochs and early stop patience of 2.
\section{Attack Algorithms}
\label{Attack}
We conduct three searching algorithms, \emph{greedy search}, \emph{beam search}, \emph{genetic algorithm} in adversarial attacks based on the real errors on NUCLE \citep{DBLP:conf/bea/DahlmeierNW13}. For \emph{beam search}, we set the beam size as 5. For \emph{genetic algorithm}, we set the population in each generation to be 60 and set the maximum number of generations to be 23\% of the corresponding sentence length. For example, if a sentence has 100 tokens, the genetic algorithm will iterate for at most 23 iterations. Algorithm \ref{greedy}, \ref{beamsearch} and \ref{genetic} are detailed descriptions of greedy attack, beam search attack, and genetic algorithm attack, respectively.

\section{Probing Model Ability in Identifying Errors}
\label{probe}
\subsection{The Sentence-level Binary Classification Task}
Table \ref{sent-lev} shows complete results for probing individual layers of ELMo, BERT, and RoBERTa across eight error types in the sentence-level binary classification task. We fix the parameters of pre-trained encoders and train a self-attention classifier for each layer to judge the binary linguistic acceptability of a sentence. We find that layer 1 of ELMo, middle layers of BERT, and top layers of RoBERTa perform the best in this evaluation.

\subsection{The Token-level Error Locating Task}
Table \ref{tok-lev} shows complete results for probing individual layers of ELMo, BERT, and RoBERTa across eight error types in the token-level. We first fix the parameters of pre-trained encoders and train a self-attention classifier for each layer to judge the binary linguistic acceptability of a sentence. Then, we extract the two positions with the highest attention weights of self-attention layers and see if error tokens are included.

\section{Case Study of Locating Error Positions}
\label{case study}
We show some examples of the token-level evaluation in section 5. We randomly select one example for each error type and visualize the attention weights of the self-attention layer upon different layers of BERT. A deeper purple under each token means the self-attention layer is putting more attention on this token.

\section{The Token-level Evaluation on Attention Heads of BERT}
\label{heads}
As mentioned in section 5. We also conduct the same token-level probing to 144 attention heads of BERT. In this experiment, the parameters in BERT are completely frozen. We observe that even in this unsupervised manner, some attention heads are still capable of precisely locating error positions. Middle layers of BERT perform the best. We further observe that some attention heads might be associated with specific types of errors. For example, head 2 in layer 9 and head 6 in layer 10 are good at capturing \texttt{SVA} and \texttt{Vform}. Both of these two errors are related to verbs.

\begin{algorithm}[t]{10cm}
    \small
    \caption{Greedy attack}
    \textbf{Input:} Original sentence $X_{ori} = \{w_1, w_2, ..., w_n\}$, ground truth prediction $Y_{ori}$, target model $F$, all confusion sets $P$, budget $b$.\\
    \textbf{Output:} Adversarial example $X_{adv}$.
    \begin{algorithmic}[1]
    \State Initialization: $X_{adv} \gets X_{ori}$
    \For{each $w_i$ in $X_{ori}$}
        \State Delete $w_i$ and compute the drop of likelihood on $Y_{ori}$
        \Statex\hspace{\algorithmicindent}to obtain the importance score of $w_i$, denoted as $S_{w_{i}}$.
        \State Apply all substitutions of $P$ to $w_{i}$. Obtain the 
        \Statex\hspace{\algorithmicindent}operation pool of $w_{i}$, denoted as $W^{sub}_i$.
    \EndFor
    \State
    \State Get the index list of $X_{ori}$ according to the decrease order of token importance: $I \gets argsort_{{w_i} \in X_{ori}}(S_{w_i})$
    \For{each $i$ in $I$}
    \State $p_{ori} \gets F(X_{adv})|_{Y = {{Y_{ori}}}}$
    \For{each $w^{'}$ in $W^{sub}_i$}
        \State Substitute $w_i$ with $w^{'}$ in $X_{adv}$ (or swap their \Statex\hspace{10mm}positions),
        \State $Y_{adv} \gets argmaxF(X_{adv})$, 
        \Statex\hspace{10mm}$p_{adv} \gets F(X_{adv})|_{Y = {{Y_{ori}}}}$
        \If {not $Y_{ori} = Y_{adv}$} \Return $X_{adv}$
        \Else 
            \If {$p_{adv} < p_{ori}$}
            \State $w_{select} \gets w^{'}$, $p_{ori} \gets p_{adv}$
            \EndIf
        \EndIf

    \EndFor
    \If{ the number of iterations exceed $b$} \Return $X_{ori}$ \EndIf 
    \State Substitute $w_i$ with $w_{select}$ in $X_{adv}$,
    \EndFor\\
    \Return $X_{ori}$
    \end{algorithmic}
\label{greedy}
\end{algorithm}
\clearpage
\begin{algorithm}[H]
    \small
    \caption{Beam search attack}
    \textbf{Input:} Original sentence $X_{ori} = \{w_1, w_2, ..., w_n\}$, ground truth prediction $Y_{ori}$, target model $F$, all confusion sets $P$, budget $b$, beam size $bm$.\\
    \textbf{Output:} Adversarial example $X_{adv}$. 
    \begin{algorithmic}[1]
    \State Initialization: $bestBeam$ $\gets$ copy $X_{ori}$ for $bm$ times.
    \For{each $w_i$ in $X_{ori}$}
        \State Delete $w_i$ and compute the drop of likelihood on $Y_{ori}$
        \Statex\hspace{\algorithmicindent}to obtain the importance score of $w_i$, denoted as $S_{w_{i}}$.
        \State Apply all substitutions of $P$ to $w_{i}$. Obtain the 
        \Statex\hspace{\algorithmicindent}operation pool of $w_{i}$, denoted as $W^{sub}_i$.
    \EndFor
    \State
    \State Get the index list of $X_{ori}$ according to the decrease order of token importance: $I \gets argsort_{{w_i} \in X_{ori}}(S_{w_i})$
     \For{each $w^{'}$ in $W^{sub}_{I[0]}$}
        \State Substitute $w_i$ with $w^{'}$ in $X_{ori}$ (or swap their \Statex\hspace{\algorithmicindent}positions)
        \State $Y_{adv} \gets argmaxF(X_{ori})$, 
        \Statex\hspace{\algorithmicindent}$p_{adv} \gets F(X_{ori})|_{Y = {{Y_{ori}}}}$
        \If {not $Y_{ori} = Y_{adv}$} \Return $X_{ori}$
        \Else 
            \State $topBeam \gets$ Record top-$bm$ examples with the 
            \Statex\hspace{10mm}lowest $p_{adv}$
        \EndIf
    \EndFor
    \State
    \State $bestBeam \gets topBeam$
    \For{each $i$ in $I / I[0]$}
    \State $p_{ori} \gets F(X_{adv})|_{Y = {{Y_{ori}}}}$
    \State $oplist \gets \{\}$
    \For{each $X_{beam}$ in $bestBeam$}
    \For{each $w^{'}$ in $W^{sub}_i$}
        \State Substitute $w_i$ with $w^{'}$ in $X_{beam}$ (or swap their \Statex\hspace{15mm}positions)
        \State $Y_{adv} \gets argmaxF(X_{beam})$, 
        \Statex\hspace{15mm}$p_{adv} \gets F(X_{beam})|_{Y = {{Y_{ori}}}}$
        \If {not $Y_{ori} = Y_{adv}$} \Return $X_{beam}$
        \Else 
            \State Add $op \gets (w^{'}, p_{adv}, X_{beam})$ to $oplist$
        \EndIf
    \EndFor
    \EndFor
    \If{ number of iterations exceed $b$} \Return $X_{ori}$ \EndIf 
    \State Select the top-$bm$ $op$s in $oplist$ with lowest $op.p_{adv}$. 
    \Statex\hspace{5mm}Update $bestBeam$ with each $op.X_{beam}$.
    \EndFor\\
    \Return $X_{ori}$
    \end{algorithmic}
\label{beamsearch}
\end{algorithm}

\begin{algorithm}[H]
    \small
    \caption{Genetic attack}
    \textbf{Input:} Original sentence $X_{ori} = \{w_1, w_2, ..., w_n\}$, ground truth prediction $Y_{ori}$, target model $F$, all confusion sets $P$, budget $b$, population size $ps$, generation size $G$.\\
    \textbf{Output:} Adversarial example $X_{adv}$.
    \begin{algorithmic}[1]
    \State Initialize the first generation with empty set: $ P^0 \gets \emptyset$.
    \For{each $w_i$ in $X_{ori}$}
        \State Apply all substitutions of $P$ to $w_{i}$. Obtain the 
        \Statex\hspace{5mm}operation pool of $w_{i}$, denoted as $W^{sub}_i$.
    \EndFor
    \For{$i = 1, 2, 3, ... , ps$}
    \State Randomly select a position $j$ and an operation from \Statex\hspace{5mm}$W^{sub}_j$ to modify $X_{ori}$. Then add to $P^0$.
    \EndFor
    \State
    \For{$g = 1, 2, 3, ... , G - 1$}
    \For{$i = 1, 2, 3, ... , ps$}
    \State $Y_{adv} \gets argmaxF(P^{g - 1}_i)$, 
    \Statex\hspace{10mm}$p_{adv} \gets F(P^{g - 1}_i)|_{Y = {{Y_{ori}}}} $
    \If{not $Y_{adv} = Y_{ori}$} \Return $P^{g - 1}_i$
    \Else
        \State $X_{elite} \gets argmin(p_{adv})$
        \State $P^{g}_1 \gets \{X_{elite}\}$
        \State $prob \gets$ Normalize sample probability with \Statex\hspace{15mm}$F(P^{g - 1}_i)$
        \For{$i = 2, 3, ... , ps$}
            \State Sample parent1 from $P^{g - 1}$ with probs 
            \Statex\hspace{20mm}$prob$
            \State Sample parent2 from $P^{g - 1}$ with probs 
            \Statex\hspace{20mm}$prob$
            \State $child \gets $ Crossover(parent1, parent2)
            \State $child_{mut} \gets $ Randomly select a position 
            \Statex\hspace{20mm}and an operation from $W^{sub}_j$ to modify \Statex\hspace{20mm}$child$
            \State $P^{g}_i \gets child_{mut}$
        \EndFor
    \EndIf
    \EndFor
    \EndFor
    \State \Return $X_{ori}$
    \end{algorithmic}
\label{genetic}
\end{algorithm}

\begin{table*}[bp]
\centering
\small
\begin{tabular}{ccccccccc}
\toprule
 &\bf{Prep} & \bf{Artordet}&\bf{Vform}&\bf{Nn}&\bf{Wchoice}&\bf{Trans}&\bf{SVA}&\bf{Worder}\cr
\toprule
ELMo, layer 0&62.6&65.0&69.6&67.7&74.5&67.5&72.1&47.6\cr
ELMo, layer 1&\textbf{90.6}&\textbf{84.7}&\textbf{87.2}&\textbf{82.9}&\textbf{83.9}&\textbf{80.6}&\textbf{93.1}&\textbf{71.2}\cr
ELMo, layer 2&84.7&77.0&79.4&79.7&82.6&74.4&89.9&68.5\cr
\toprule
 BERT, layer 0 & 62.5 & 60.8 & 67.4 & 64.6 & 73.9 & 69.5 & 70.3 & 48.2 \cr
 BERT, layer 1 &68.0 & 63.4 & 69.3 & 70.3 & 75.0 & 71.5 & 78.4 & 52.2 \cr
 BERT, layer 2 &74.4 & 67.0 & 75.3 & 74.8 & 76.7 & 73.1 & 84.4 & 62.0 \cr
 BERT, layer 3 &80.5 & 75.0 & 83.4 & 73.7 & 78.5 & 76.3 & 89.2 & 69.8 \cr
 BERT, layer 4 &82.7 & 80.7 & 83.6 & 77.7 & 82.6 & 79.6 & 90.6 & 72.4 \cr
 BERT, layer 5 &85.2 & 83.8 & 85.4 & 84.3 & 84.5 & 81.8 & 91.7 & 71.9 \cr
 BERT, layer 6 &88.2 & 86.6 & 85.8 & 86.7 & 84.5 & 82.6 & 90.9 & 73.4 \cr
 BERT, layer 7 &91.3 & 88.1 & 90.2 & 86.5 & \textbf{86.9} & 83.9 & \textbf{95.3} & 73.4 \cr
 BERT, layer 8 &\textbf{92.5} & \textbf{88.3} & \textbf{91.4} & \textbf{88.4} & 86.3 & \textbf{85.5} & 94.5 & \textbf{73.8} \cr
 BERT, layer 9 &91.4 & 86.3 & 89.9 & 87.4 & 85.6 & 84.9 & 94.4 & 72.4 \cr
 BERT, layer 10 &90.8 & 87.4 & 88.2 & 87.0 & 86.1 & 84.8 & 94.9 & 71.8 \cr
 BERT, layer 11 &90.0 & 84.9 & 88.1 & 86.6 & 85.6 & 84.3 & 94.2 & 69.5 \cr
 BERT, layer 12 &88.4& 85.6& 88.1& 84.3& 84.0& 82.6& 93.3& 68.1\cr
\toprule
 RoBERTa, layer 0 & 61.9 & 65.9 & 69.7 & 67.1 & 75.1 & 69.1 & 68.3 & 50.9 \cr
 RoBERTa, layer 1 & 78.3 & 74.7 & 84.6 & 77.6 & 80.2 & 75.9 & 88.4 & 67.8 \cr
 RoBERTa, layer 2 & 85.2 & 79.4 & 88.7 & 83.0 & 83.3 & 78.8 & 90.9 & 71.8 \cr
 RoBERTa, layer 3 & 89.3 & 85.7 & 90.6 & 86.9 & 87.0 & 84.1 & 94.3 & 72.6 \cr
 RoBERTa, layer 4 & 90.2 & 88.7 & 91.8 & 88.7 & 86.2 & 86.4 & 94.5 & 74.5 \cr
 RoBERTa, layer 5 & 91.4 & 89.1 & 92.9 & 90.5 & 89.0 & 87.1 & 95.5 & 74.5 \cr
 RoBERTa, layer 6 & 93.4 & 91.3 & 91.9 & 91.4 & 88.9 & 86.8 & 95.0 & 75.3 \cr
 RoBERTa, layer 7 & 93.9 & 90.5 & 91.8 & 90.4 & 88.2 & 86.9 & 94.6 & 74.7 \cr
 RoBERTa, layer 8 & 93.9 & 91.1 & \textbf{93.4} & 92.3 & 88.0 & 87.2 & 94.4 & 75.9 \cr
 RoBERTa, layer 9 & 94.3 & 90.6 & 92.5 & 92.1 & 89.4 & 88.0 & \textbf{95.7} & 74.7 \cr
 RoBERTa, layer 10 & 94.4 & \textbf{92.0} & 93.3 & \textbf{92.3} & \textbf{89.9} & 88.1 & 95.0 & 75.1 \cr
 RoBERTa, layer 11 & \textbf{95.3} & 91.5 & 93.3 & 89.4 & 88.8 & \textbf{88.2} & 95.2 & \textbf{76.0} \cr
 RoBERTa, layer 12 & 94.5& 91.1& 92.7& 88.3& 87.3& 87.9& 95.3& 74.8\cr
\bottomrule\hline
\end{tabular}
\caption{Results of the accuracy on the binary linguistic acceptability probing task for individual layers of ELMo, BERT, and RoBERTa.}

\label{sent-lev}
\end{table*}
\begin{table*}[bp]
\centering
\small
\begin{tabular}{ccccccccc}
\toprule
 &\bf{Prep} & \bf{Artordet}&\bf{Vform}&\bf{Nn}&\bf{Wchoice}&\bf{Trans}&\bf{SVA}&\bf{Worder}\cr
\toprule
ELMo, layer 0 & 23.2&14.3&22.3&9.8&21.8&10.2&18.4&29.6\cr
ELMo, layer 1 &56.5&\textbf{42.6}&51.8&82.0&72.0&\textbf{69.4}&30.6&55.1\cr
ELMo, layer 2 &\textbf{68.0}&34.2&\textbf{55.4}&\textbf{85.9}&\textbf{73.0}&42.8&\textbf{49.2}&\textbf{62.7}\cr
\toprule
 BERT, layer 0&24.1 & 39.1 & 66.7 & 58.7 & 62.3 & 56.4 & 63.6 & 17.5 \cr
 BERT, layer 1&56.6 & 33.9 & 66.9 & 59.3 & 69.4 & 71.1 & 54.4 & 13.1 \cr
 BERT, layer 2&58.7 & 27.4 & 75.8 & 58.4 & 76.3 & 83.3 & 60.0 & 34.1 \cr
 BERT, layer 3&64.5 & 55.2 & 56.2 & 62.4 & 79.3 & 83.0 & 64.2 & 67.8 \cr
 BERT, layer 4&68.9 & 54.1 & 69.2 & 62.9 & 81.7 & 66.0 & 67.3 & 59.7 \cr
 BERT, layer 5&67.4 & 52.4 & 76.9 & 60.8 & 83.8 & 80.7 & 62.2 & 62.3 \cr
 BERT, layer 6&68.2 & 51.5 & 76.5 & 58.7 & 84.9 &\textbf{83.9} & 71.7 & 66.9 \cr
 BERT, layer 7&70.4 &\textbf{52.3} & 93.0 & 61.8 & 82.8 & 81.9 & 61.3 & 61.2 \cr
 BERT, layer 8&69.9 & 51.7 & 93.0 & 65.4 & 80.2 & 80.2 & 60.9 &\textbf{63.9} \cr
BERT, layer 9&\textbf{71.7} & 48.0 & 91.6 &\textbf{85.3} &\textbf{84.9} & 79.6 & 59.6 & 62.2 \cr
 BERT, layer 10&70.7 & 50.4 & 90.5 & 80.5 & 82.3 & 78.2 & 92.4 & 58.7 \cr
 BERT, layer 11&70.1 & 49.2 &\textbf{96.3} & 80.5 & 81.0 & 80.7 & 90.5 & 60.3 \cr
 BERT, layer 12&71.4& 50.5& 86.7& 79.8& 79.1& 81.6&\textbf{93.2}& 58.8\cr
\toprule
 RoBERTa, layer 0 & 44.8&26.5&74.8&62.8& 71.3&71.1&61.7&14.3\cr
 RoBERTa, layer 1 & 68.3& 12.1& 90.7& 62.5& 80.9& 75.9& 93.5& 48.9\cr
 RoBERTa, layer 2 & 69.9& 35.3& 71.0& 61.9& 83.9& 84.1& 60.5& 58.2\cr
 RoBERTa, layer 3 & 71.9& 54.4& 92.2& 60.7& 85.5& 84.4& \textbf{96.2}& 59.3\cr
 RoBERTa, layer 4 & 71.2& 48.9&92.0& 83.3&85.6&\textbf{85.3}& 95.9&60.8\cr
 RoBERTa, layer 5 & \textbf{71.9}& \textbf{53.6}&92.5& 84.9&\textbf{88.5}&83.9&95.3&61.2\cr
 RoBERTa, layer 6 & 70.2& 52.9&92.5&87.0&87.3&83.9&95.7&59.0\cr
 RoBERTa, layer 7 & 70.6&50.6 &92.1&87.8&87.2&83.9& 94.8&58.4\cr
 RoBERTa, layer 8 & 71.6& 51.5&92.2&\textbf{89.5}&87.0&79.6&95.2&58.8\cr
 RoBERTa, layer 9 & 71.3& 53.2&91.9&87.7&86.7&81.3&95.8&61.1\cr
 RoBERTa, layer 10 & 69.6&50.3&\textbf{92.8}&86.8&87.1&78.8&96.0&\textbf{64.2}\cr
 RoBERTa, layer 11 & 69.3&49.6&92.7&88.4&86.5&75.6&95.5&62.0\cr
 RoBERTa, layer 12 & 69.6&48.9&90.1&86.8&84.9&79.6&94.1&62.8\cr
\bottomrule\hline
\end{tabular}
\caption{Results of the accuracy on locating error positions for individual layers of ELMo, BERT, and RoBERTa.}
\label{tok-lev}
\end{table*}
\clearpage

\begin{figure*}[!htb]
\centering
\includegraphics[width=70 mm, height=48mm]{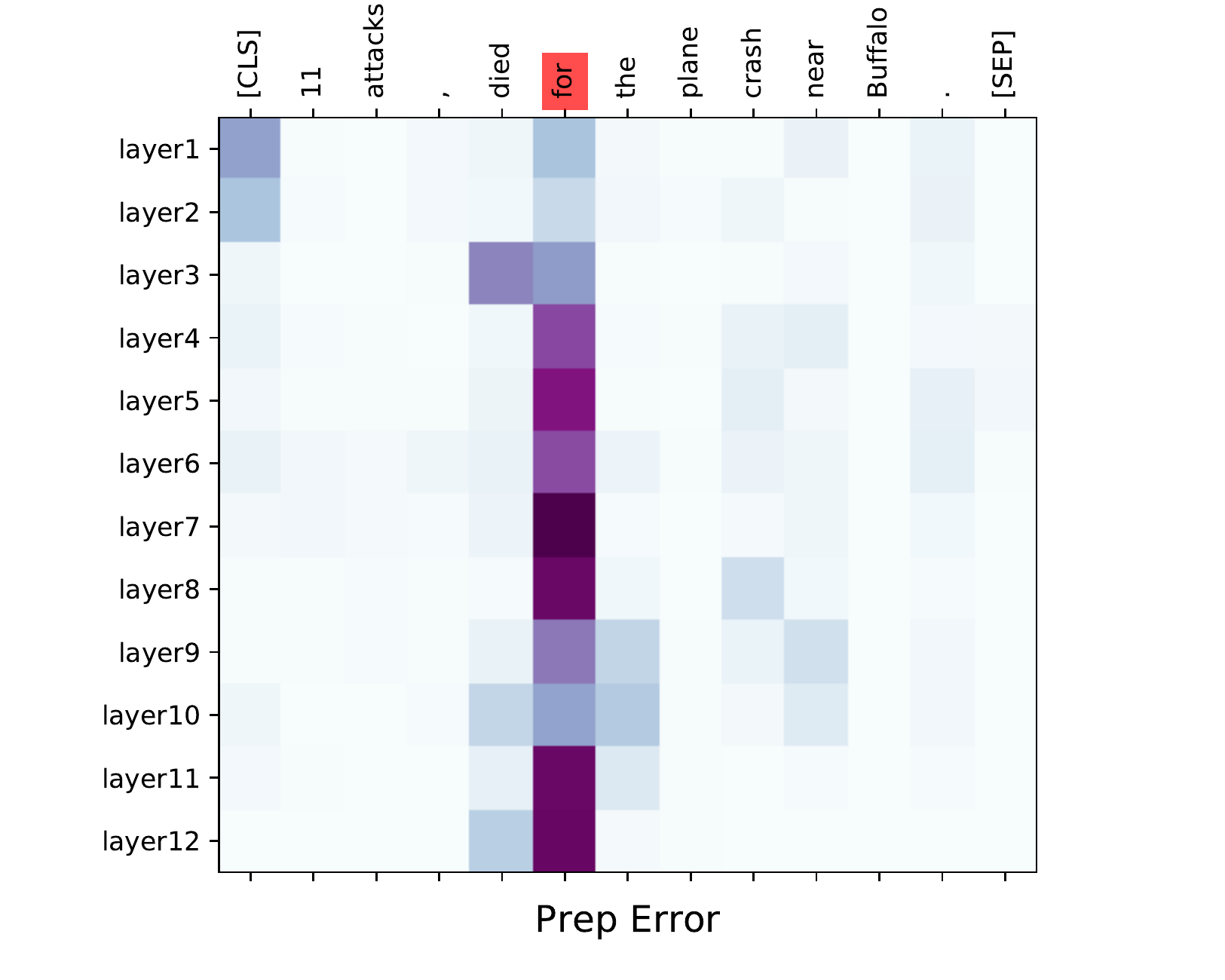}
\includegraphics[width=70 mm, height=48mm=]{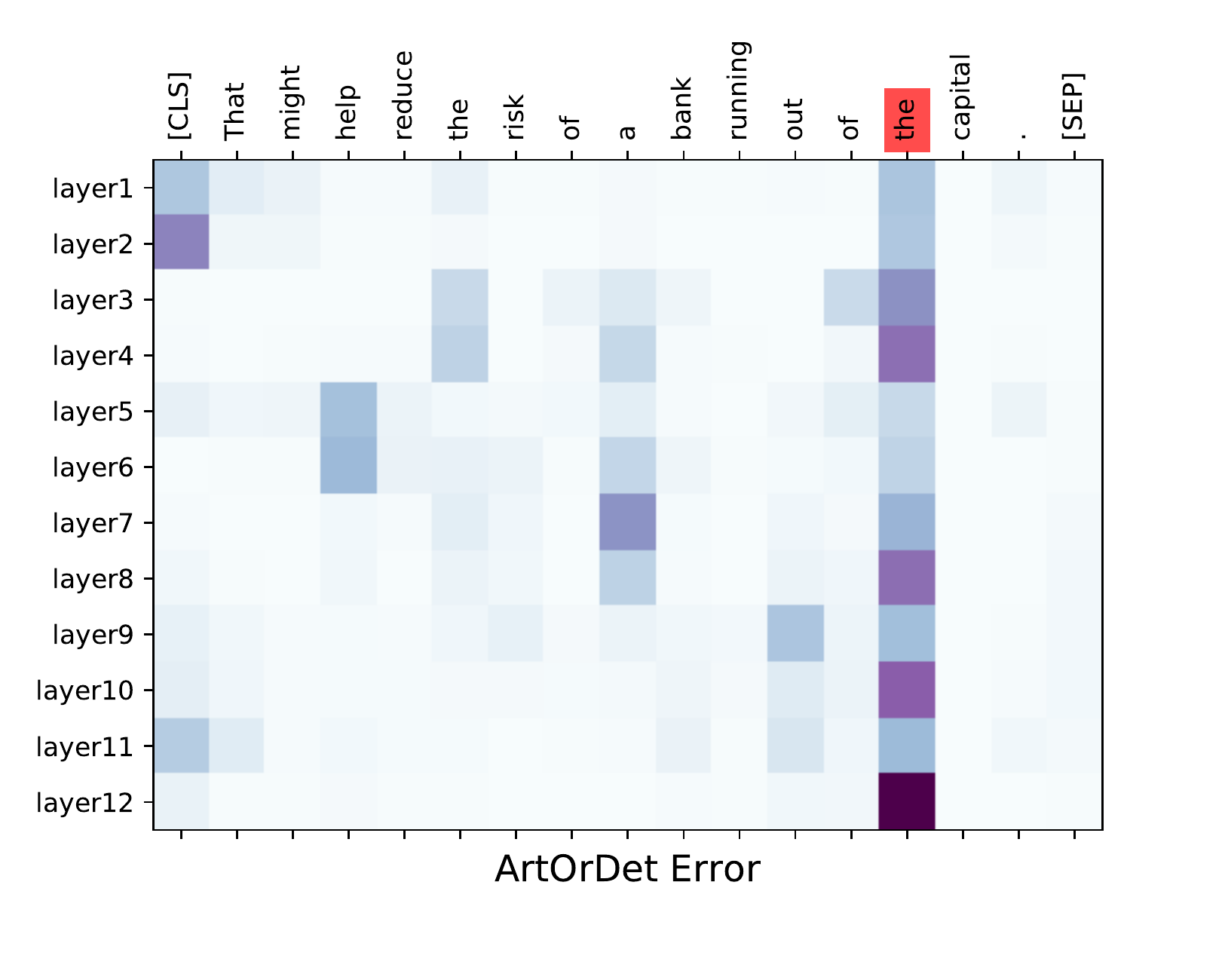}
\includegraphics[width=70 mm, height=48mm=]{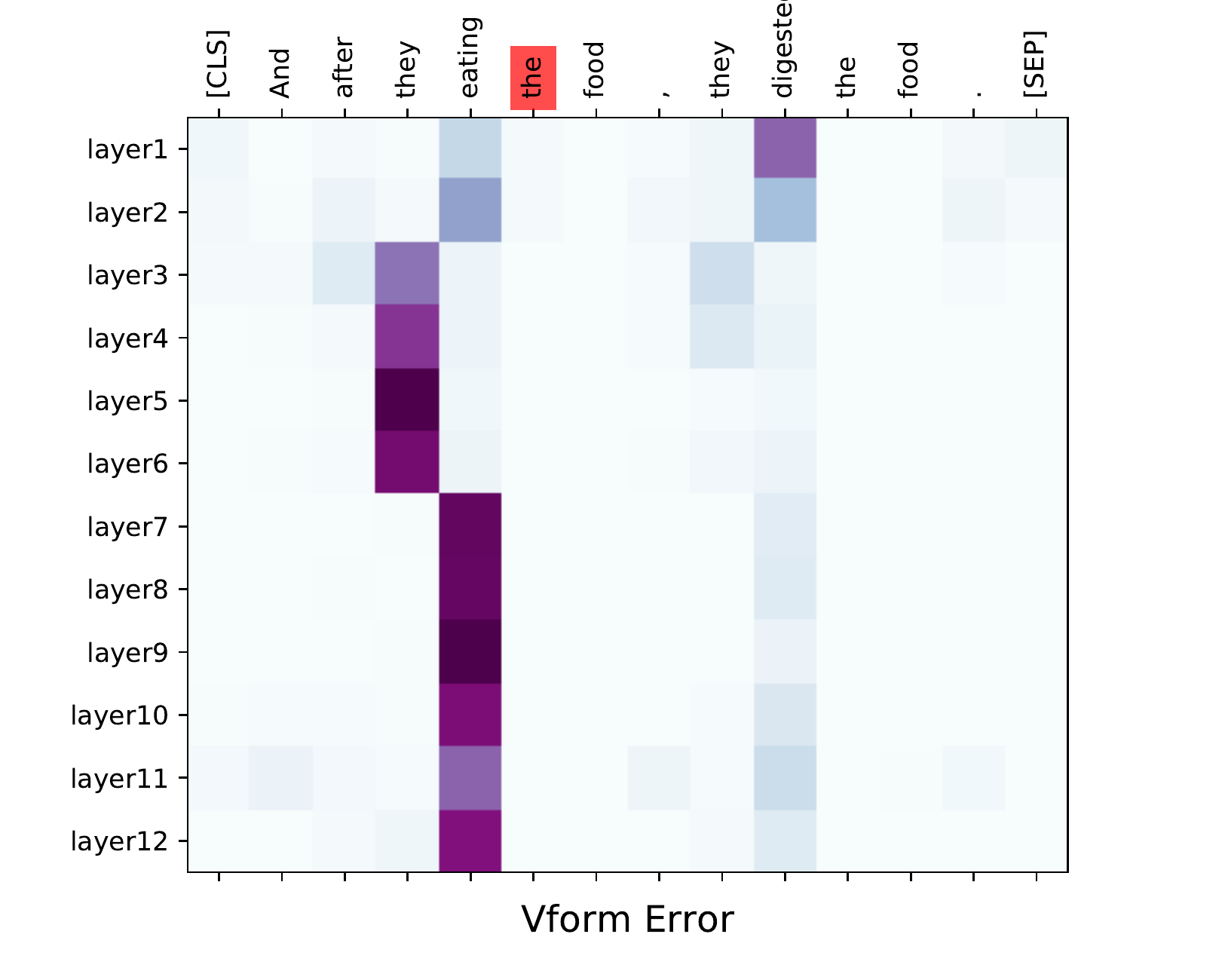}
\includegraphics[width=70 mm, height=48mm=]{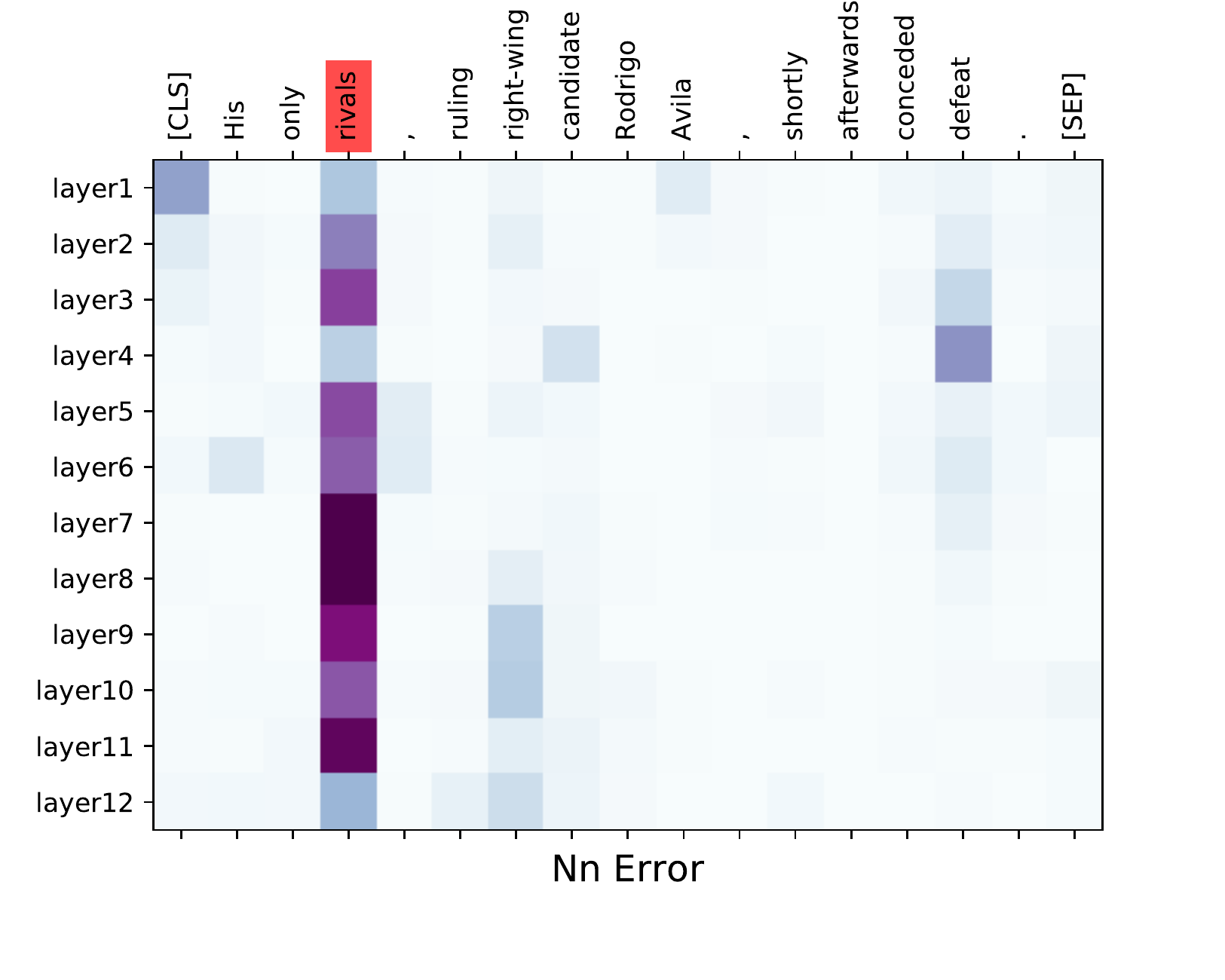}
\includegraphics[width=70 mm, height=48mm]{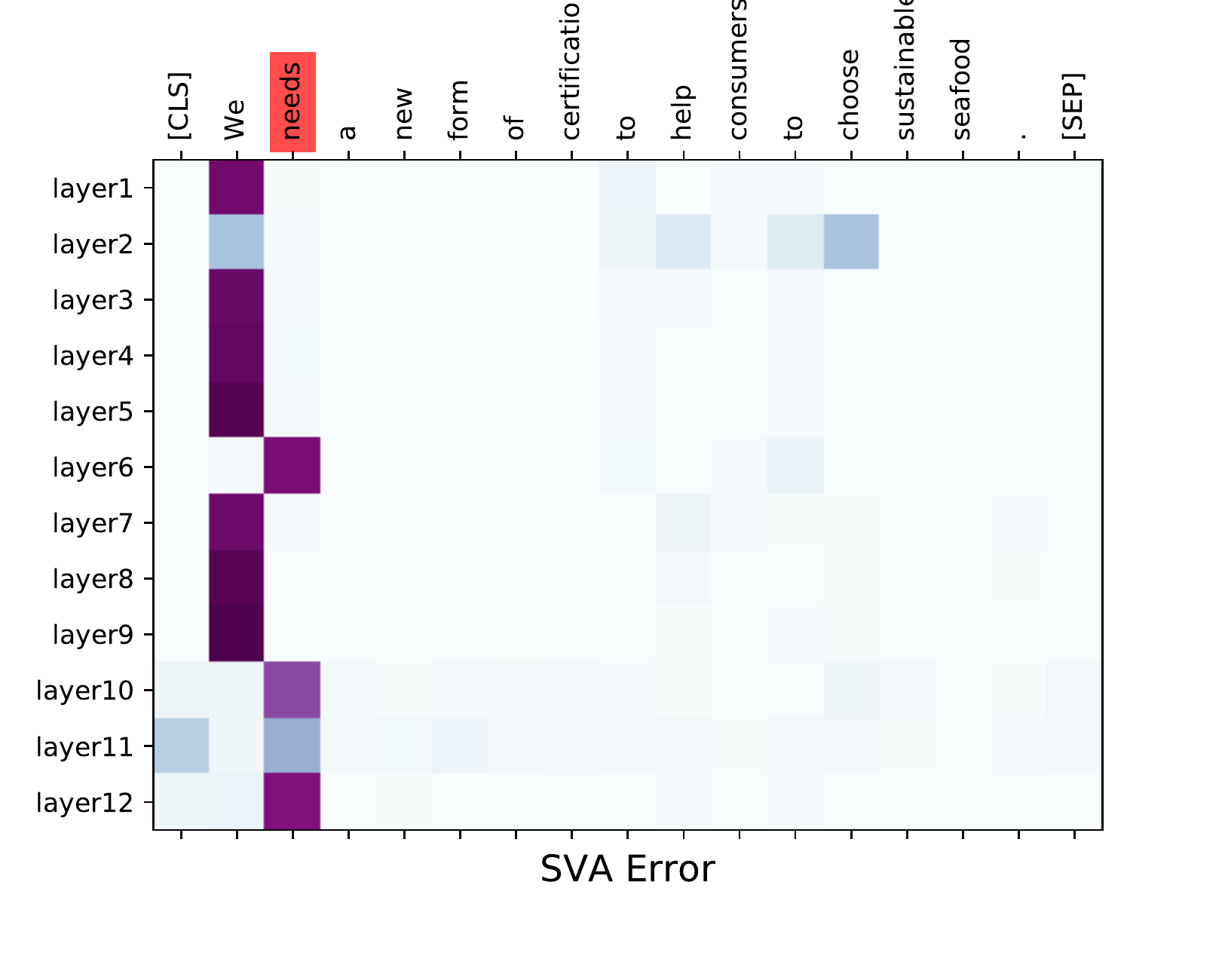}
\includegraphics[width=70 mm, height=48mm]{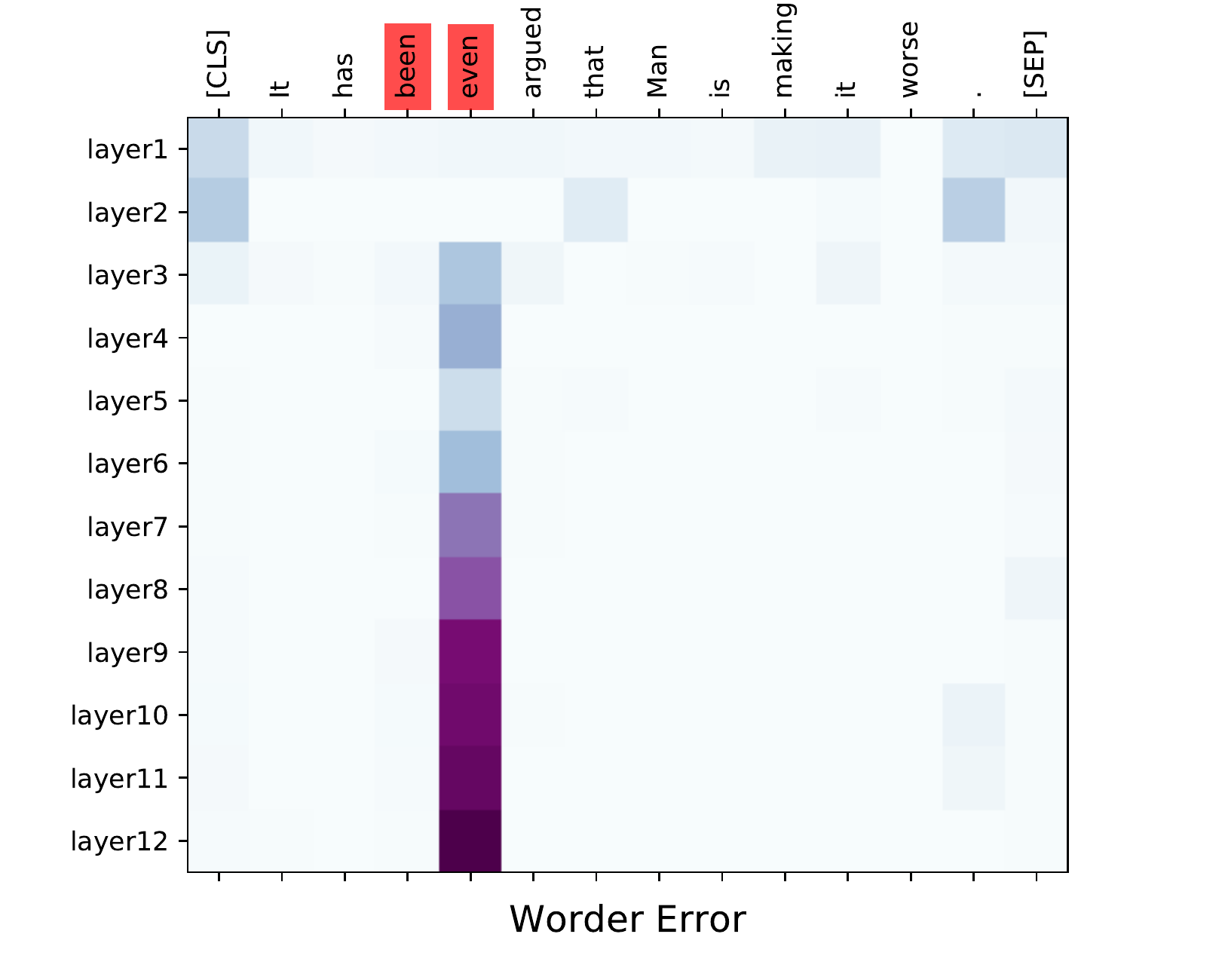}
\includegraphics[width=70 mm, height=48mm]{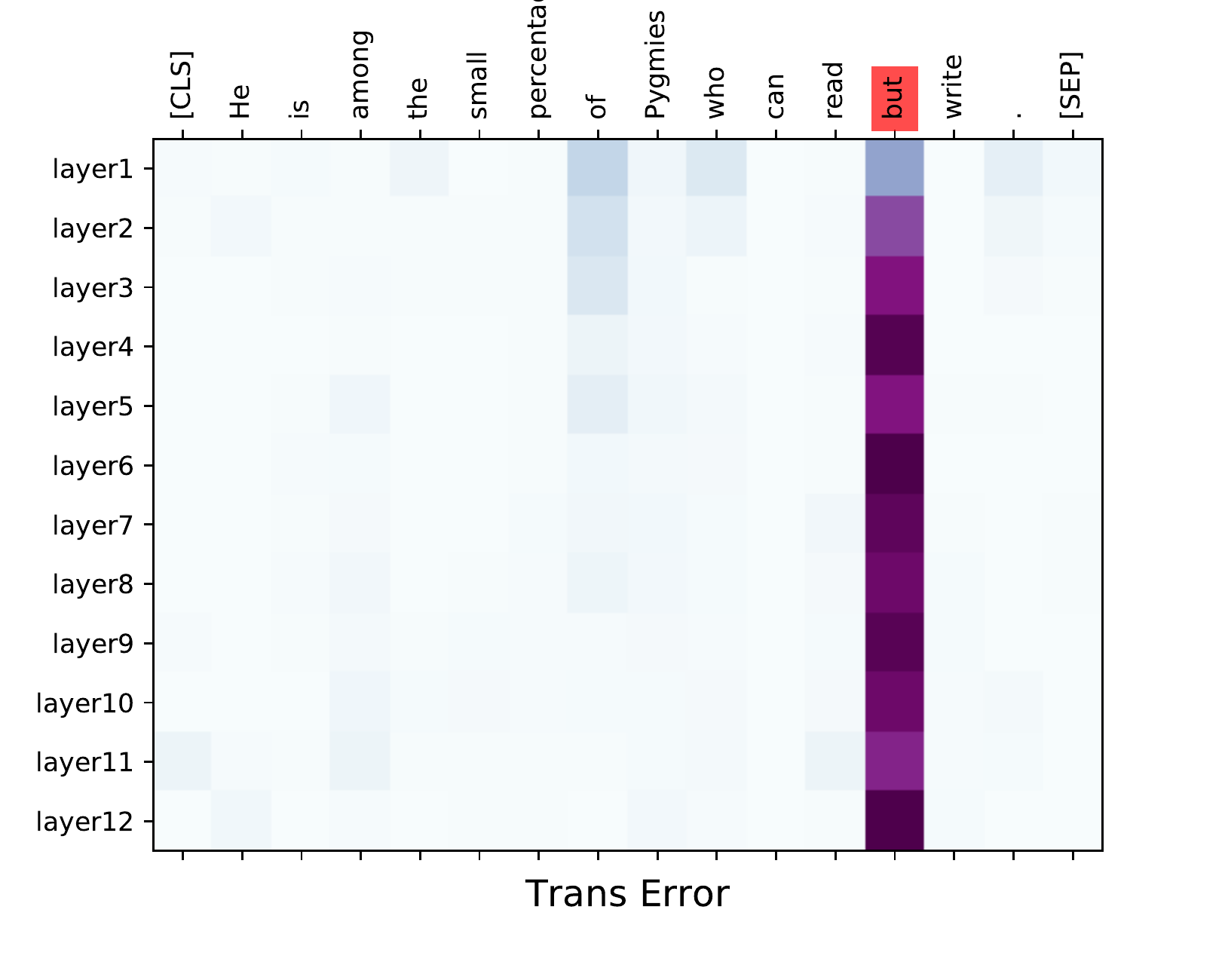}
\includegraphics[width=70 mm, height=48mm]{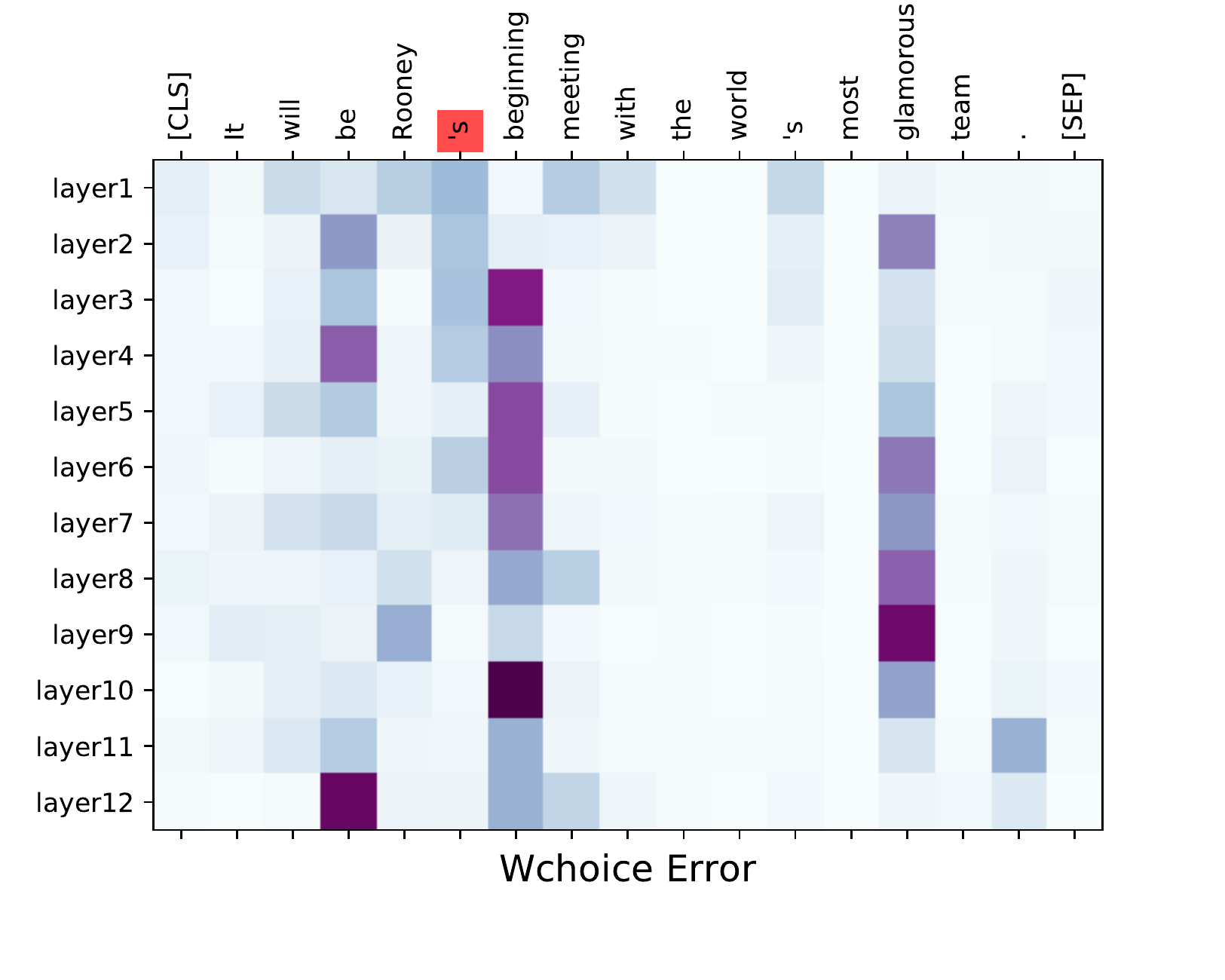}
\caption{Visualization of attention weights of self-attention layers. A figure represents a sentence with a specific error type. Errors in a sentence are highlighted in red. Each column represents one layer of BERT that the self-attention layer is build upon.}
\end{figure*}

\begin{figure*}[!htb]
    \centering
    \includegraphics[width=78 mm, height=40mm]{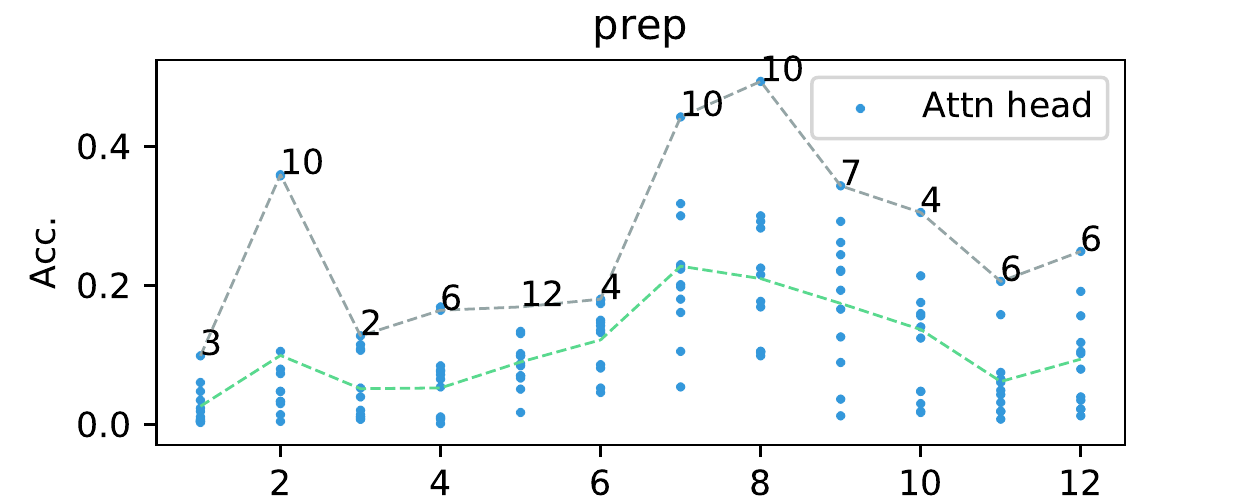}
    \includegraphics[width=78 mm, height=40mm]{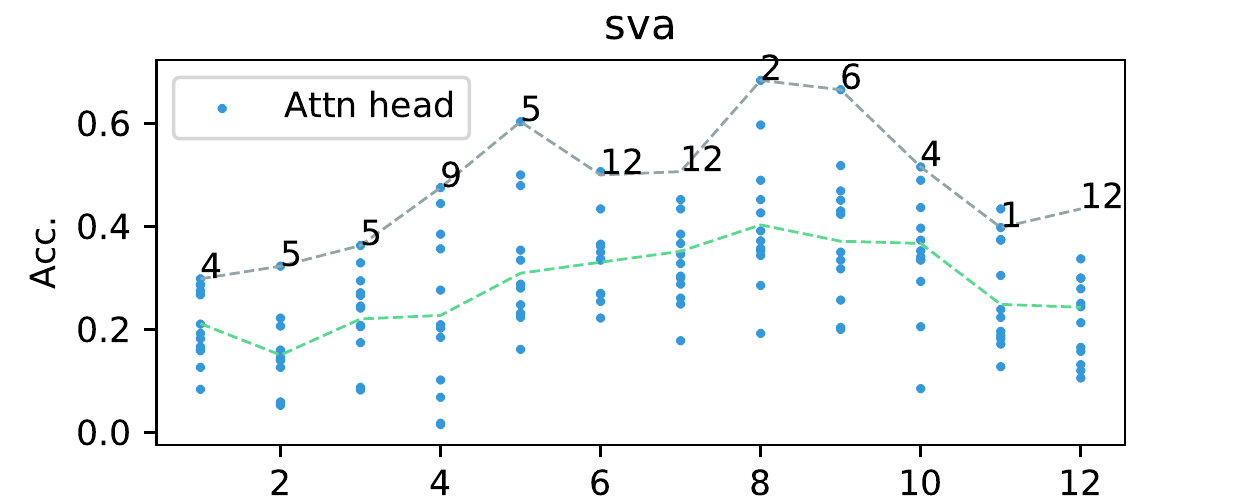}
    \includegraphics[width=78 mm, height=40mm]{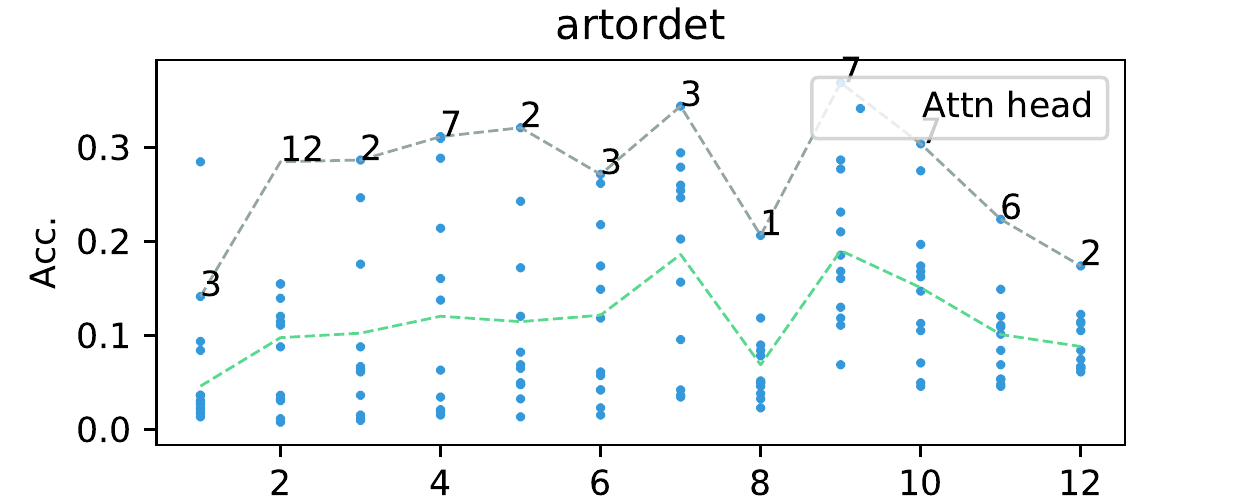}
    \includegraphics[width=78 mm, height=40mm]{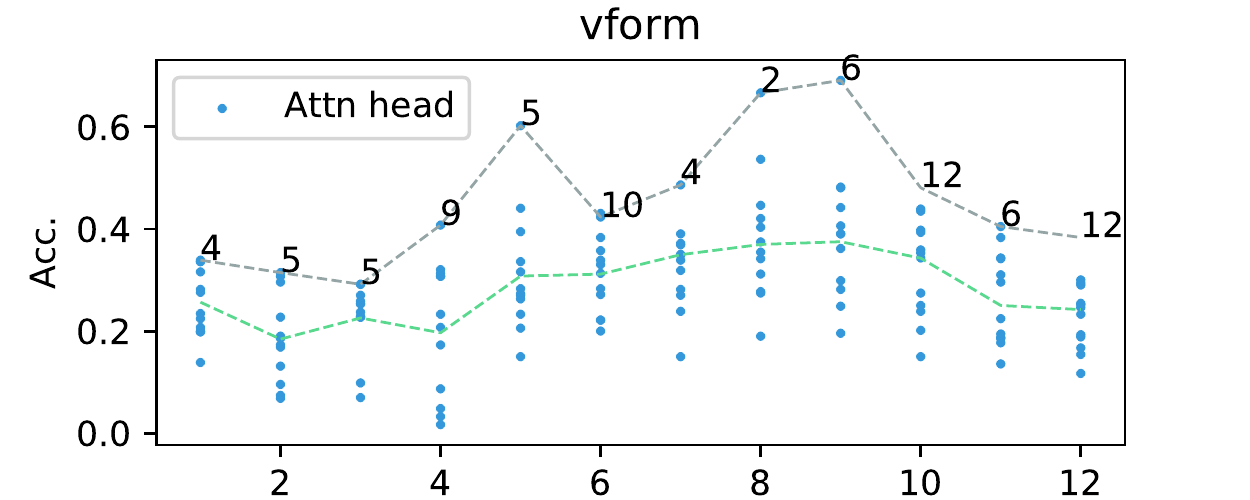}
    \includegraphics[width=78 mm, height=40mm]{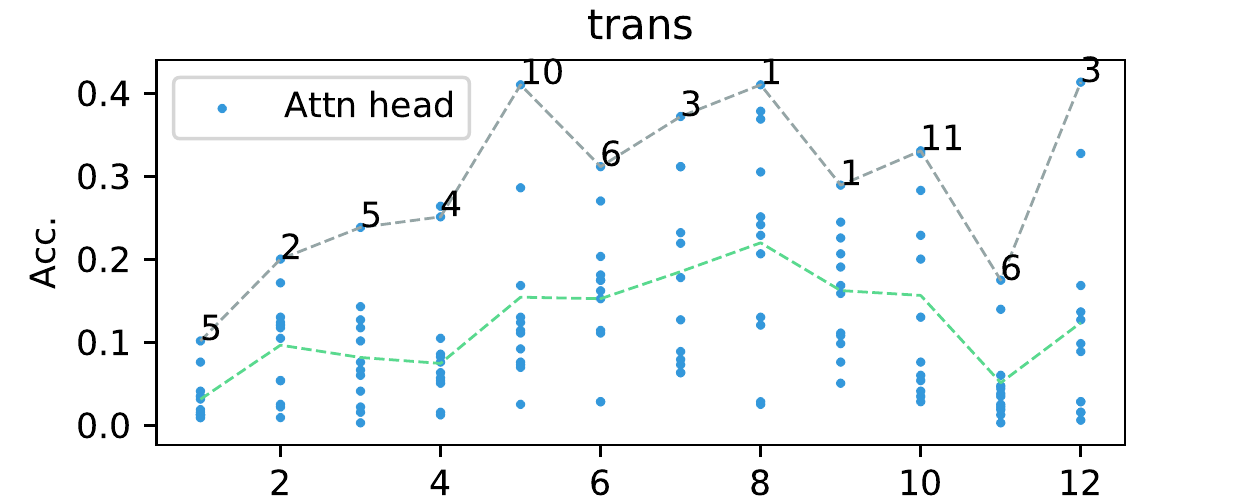}
    \includegraphics[width=78 mm, height=40mm]{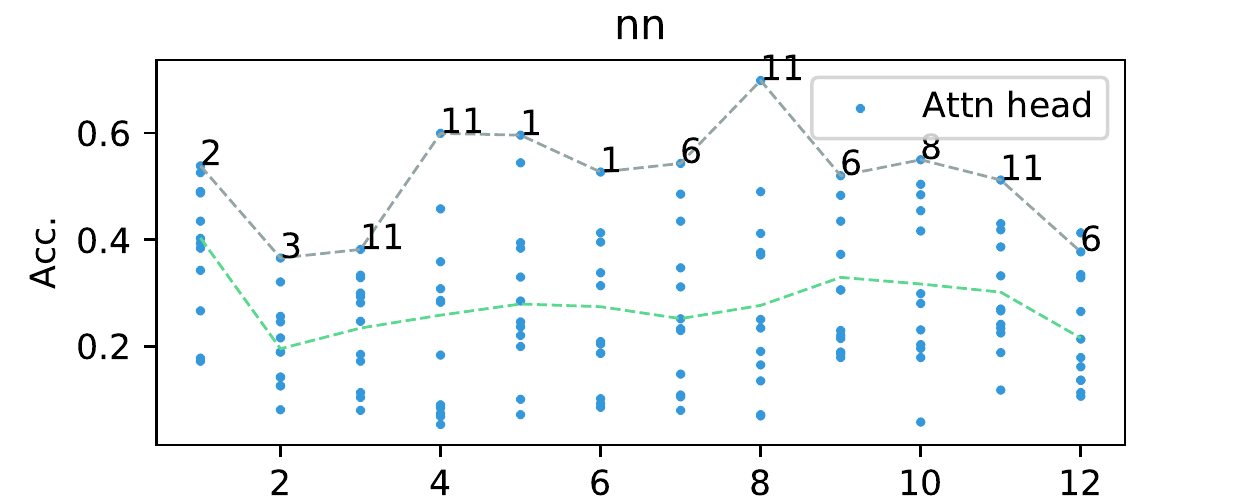}
    \includegraphics[width=78 mm, height=40mm]{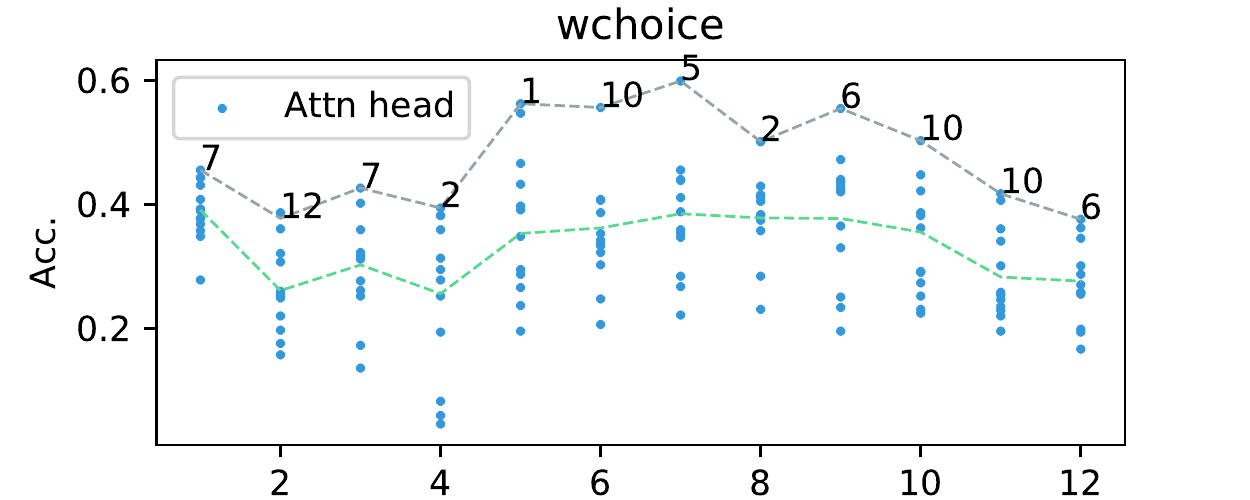}
    \includegraphics[width=78 mm, height=40mm]{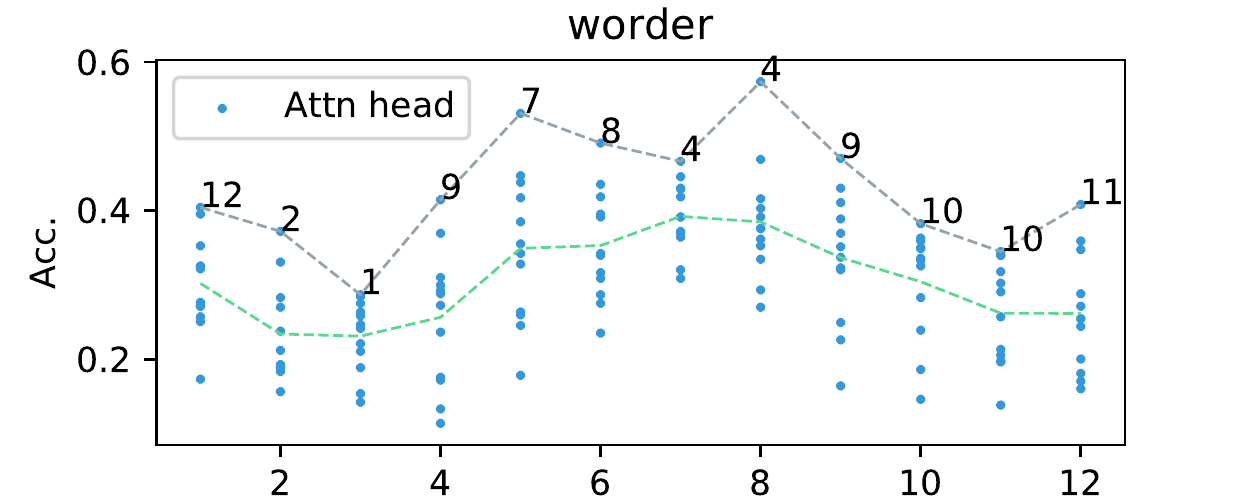}
\caption{Visualization for each attention head of BERT for locating each type of error. A point in the figure represents the performance of an attention head. The grey line on the top represents the best performing head in each layer (annotated with its number). The green line in the middle represents the average performance of all heads in this layer.}
\end{figure*}
\onecolumn
\clearpage
    
\section{Human Evaluation and Samples of Successful Attack Example}
\label{example}
%MRPC
\subsection{Human Evaluation}
To evaluate the validation of synthesized examples, we randomly select 200 attacked examples (greedy attack on BERT) from the development set of MNLI-matched and SST-2 (100 examples for each). We ask two of our authors to make predictions on both attacked and original examples and see if the predictions will change after the attack in human perspectives. We find that humans make the same predictions to 87.5\% of MNLI examples (averaged between two annotators, 89\% and 86\% respectively) and 88\% of SST-2 examples (both two annotators give 88\%). In comparison, the original annotation agreement for MNLI is around 92\%. This shows that though our algorithm is not perfect, it keeps a majority of true labels unchanged.

\subsection{Successful Examples for MRPC}
Table \ref{MRPC} shows some successful adversarial attack examples on MRPC dataset.
\begin{table*}[!htb]
\centering
\begin{tabularx}{1.0\textwidth}{ >{\raggedright\arraybackslash}X}
\toprule
\textbf{Original:}\\
\toprule
\textbf{Sentence 1:} The association said 28.2 million DVDs were rented in the \colorbox{blue!20}{week} that ended \colorbox{blue!20}{June} 15, \colorbox{blue!20}{compared} with 27.3 million VHS cassettes.\\
\textbf{Sentence 2:} The Video Software Dealers Association said 28.2 million DVDs were rented out last week, compared to 27.3 million VHS cassettes.\\
\toprule
\textbf{Changed (beam search):}\\
\toprule
\textbf{Sentence 1:} The association said 28.2 million DVDs were rented in the \colorbox{red!20}{workweek} that ended \colorbox{red!20}{Junes} 15, \colorbox{red!20}{liken} with 27.3 million VHS cassettes.\\
\textbf{Sentence 2:} The Video Software Dealers Association said 28.2 million DVDs were rented out last week, compared to 27.3 million VHS cassettes.\\
\toprule
\end{tabularx}

\begin{tabularx}{1.0\textwidth}{ >{\raggedright\arraybackslash}X}
\toprule
\textbf{Original:}\\
\toprule
\textbf{Sentence 1:} A European Union \colorbox{blue!20}{spokesman} \colorbox{blue!20}{said} the Commission was consulting EU member states "with a view to taking appropriate action if necessary" on the matter.\\
\textbf{Sentence 2:} Laos's second most important export destination - said it was consulting EU member states ''with a view to taking appropriate action if necessary'' on the matter.\\
\toprule
\textbf{Changed (beam search):}\\
\toprule
\textbf{Sentence 1:} A European Union \colorbox{red!20}{spokesmen} \colorbox{red!20}{saying} the Commission was consulting EU member states "with a view to taking appropriate action if necessary" under the matter .\\
\textbf{Sentence 2:} Laos's second most important export destination - said it was consulting EU member states ''with a view to taking appropriate action if necessary'' on the matter.\\
\toprule
\end{tabularx}
\begin{tabularx}{1.0\textwidth}{ >{\raggedright\arraybackslash}X}
\toprule
\textbf{Original:}\\
\toprule
\textbf{Sentence 1:} Morrill's wife, Ellie, sobbed \colorbox{blue!20}{and} hugged Bondeson's sister-in-law during the service.\\
\textbf{Sentence 2:} At the service Morrill's widow, Ellie, sobbed and hugged Bondeson's sister-in-law as people consoled her.\\
\toprule
\textbf{Changed (genetic):}\\
\toprule
\textbf{Sentence 1:} Morrill's wife, Ellie, sobbed \colorbox{red!20}{as} hugged Bondeson's sister-in-law during the service.\\
\textbf{Sentence 2:} At the service Morrill's widow, Ellie, sobbed and hugged Bondeson's sister-in-law as people consoled her.\\
\bottomrule\toprule
\end{tabularx}
\caption{Successful attack examples against pre-trained encoders on MRPC. Modified words are highlighted.}
\label{MRPC}
\end{table*}

%QNLI
\clearpage
\subsection{Successful Examples for QNLI}
Table \ref{QNLI} shows some successful adversarial attack examples on QNLI dataset.

\begin{table}[H]
\centering
\begin{tabularx}{1.0\textwidth}{ >{\raggedright\arraybackslash}X}
\toprule
\textbf{Original:}\\
\toprule
\textbf{Question:} What are the most active parts of ctenophora?\\
\textbf{Answer:} These branch through the mesoglea to the most \colorbox{blue!20}{active} parts of the animal: the mouth and pharynx; the roots of the tentacles, if present; all along the underside of each comb row; and four branches round the sensory complex at the far end from the mouth – two of these four branches terminate in anal pores.\\
\toprule
\textbf{Changed (greedy):}\\
\toprule
\textbf{Question:} What are the most active parts of ctenophora?\\
\textbf{Answer:} These branch through the mesoglea to the most \colorbox{red!20}{alive} parts of the animal: the mouth and pharynx; the roots of the tentacles, if present; all along the underside of each comb row; and four branches round the sensory complex at the far end from the mouth – two of these four branches terminate in anal pores.\\
\toprule
\end{tabularx}
\begin{tabularx}{1.0\textwidth}{ >{\raggedright\arraybackslash}X}
\toprule
\textbf{Original:}\\
\toprule
\textbf{Question:} What type of flower is sought on Midsummer's Eve?\\
\textbf{Answer:} Each Midsummer’s Eve, apart from the official floating of wreaths, jumping over fires, \colorbox{blue!20}{looking} \colorbox{blue!20}{for} the fern flower, there are musical performances, dignitaries' speeches, fairs and fireworks by the river bank.\\
\toprule
\textbf{Changed (greedy):}\\
\toprule
\textbf{Question:} What type of flower is sought on Midsummer's Eve?\\
\textbf{Answer:} Each Midsummer’s Eve, apart from the official floating of wreaths, jumping over fires, \colorbox{red!20}{seem} \colorbox{red!20}{at} the fern flower, there are musical performances, dignitaries' speeches, fairs and fireworks by the river bank.\\
\toprule
\end{tabularx}
\caption{Successful attack examples against pre-trained encoders on QNLI. Modified words are highlighted.}
\label{QNLI}
\end{table}

%MNLI
\subsection{Successful Examples for MNLI}
Table \ref{MNLI} shows some successful adversarial attack examples on MNLI dataset.
\begin{table}[H]
\centering
\begin{tabularx}{1.0\textwidth}{ >{\raggedright\arraybackslash}X}
\toprule
\textbf{Original:}\\
\toprule
\textbf{Premise:} There are no shares of a stock that might someday come back, just piles of options as worthless as those shares of Cook's American Business Alliance.\\
\textbf{Hypothesis:} Those shares of stocks \colorbox{blue!20}{will never} come back.\\
\toprule
\textbf{Changed (greedy):}\\
\toprule
\textbf{Premise:} There are no shares of a stock that might someday come back, just piles of options as worthless as those shares of Cook's American Business Alliance.\\
\textbf{Hypothesis:} Those shares of stocks \colorbox{red!20}{never will} come back.\\
\toprule
\end{tabularx}

\caption{Successful attack examples against pre-trained encoders on MNLI. Modified words are highlighted.}
\label{MNLI}
\end{table}

%SST-2
\subsection{Successful Examples for SST-2}
Table \ref{SST-2} shows some successful adversarial attack examples on SST-2 dataset.
\begin{table}[H]
\centering

\begin{tabularx}{1.0\textwidth}{ >{\raggedright\arraybackslash}X}
\toprule
\textbf{Original:}\\
\toprule
pumpkin means to be an outrageous dark satire on fraternity life, but its ambitions far \colorbox{blue!20}{exceeding} the abilities of writer adam larson broder and his co-director, tony r. abrams, in their feature debut .\\

\toprule
\textbf{Changed (beam search):}\\
\toprule
pumpkin means to be an outrageous dark satire on fraternity life , but its ambitions far \colorbox{red!20}{exceed} the abilities of writer adam larson broder and his co-director, tony r. abrams, in their feature debut.\\

\toprule
\end{tabularx}

\begin{tabularx}{1.0\textwidth}{ >{\raggedright\arraybackslash}X}
\toprule
\textbf{Original:}\\
\toprule
nothing's at stake, just a twisty double-cross you can smell a mile away -- still, the derivative nine queens is lots of \colorbox{blue!20}{fun}.\\

\toprule
\textbf{Changed (beam search):}\\
\toprule
nothing's at stake, just a twisty double-cross you can smell a mile away -- still, the derivative nine queens is lots of \colorbox{red!20}{merriment}.\\

\toprule
\end{tabularx}

\begin{tabularx}{1.0\textwidth}{ >{\raggedright\arraybackslash}X}
\toprule
\textbf{Original:}\\
\toprule
it's of the quality of a lesser harrison ford movie - six days, seven nights, maybe, or that \colorbox{blue!20}{direful} sabrina remake.\\

\toprule
\textbf{Changed (beam search):}\\
\toprule
it's of the quality of a lesser harrison ford movie - six days, seven nights, maybe, or that \colorbox{red!20}{dreadful} sabrina remake.\\

\toprule
\end{tabularx}

\caption{Successful attack examples against pre-trained encoders on SST-2. Modified words are highlighted.}
\label{SST-2}
\end{table}

% NER
\subsection{Successful Examples for NER}
Table \ref{NER} shows some successful adversarial attack examples on CoNLL2003-NER dataset.
\begin{table*}[!htb]
\centering
\begin{tabularx}{1.0\textwidth}{ >{\raggedright\arraybackslash}X}
\toprule
\textbf{Original:}\\
\toprule
as well as one-day matches \colorbox{blue!20}{against} the Minor Countries and Scotland. \\

\toprule
\textbf{Changed (greedy):}\\
\toprule
as well as one-day matches \colorbox{red!20}{of} the Minor Countries and Scotland. \\

\toprule
\end{tabularx}

\begin{tabularx}{1.0\textwidth}{ >{\raggedright\arraybackslash}X}
\toprule
\textbf{Original:}\\
\toprule
The August Chicago NAPM rose 8.8 points to 60.0, its highest level since 62.6 in February 1995 and the largest monthly \colorbox{blue!20}{rise} since December 1993. \\
\toprule
\textbf{Changed (genetic):}\\
\toprule
The August Chicago NAPM rose 8.8 points to 60.0, its highest level since 62.6 in February 1995 and the largest monthly \colorbox{red!20}{rises} since December 1993. \\

\toprule
\end{tabularx}

\caption{Successful attack examples against pre-trained encoders on CoNLL2003-NER. Modified words are highlighted.}
\label{NER}
\end{table*}

\clearpage

\end{document}